\documentclass[conference]{IEEEtran}
\IEEEoverridecommandlockouts
\usepackage{cite}
\usepackage{amsmath,amssymb,amsfonts}
\usepackage{algorithmic}
\usepackage{graphicx}
\usepackage{textcomp}
\usepackage[table]{xcolor}
\usepackage{tabularx}
\usepackage{enumitem}%
\usepackage{diagbox}
\usepackage{caption}
\usepackage{subcaption}

\renewcommand{\texttt}[1]{%
  \begingroup
  \ttfamily
  \begingroup\lccode`~=`/\lowercase{\endgroup\def~}{/\discretionary{}{}{}}%
  \begingroup\lccode`~=`[\lowercase{\endgroup\def~}{[\discretionary{}{}{}}%
  \begingroup\lccode`~=`.\lowercase{\endgroup\def~}{.\discretionary{}{}{}}%
  \catcode`/=\active\catcode`[=\active\catcode`.=\active
  \scantokens{#1\noexpand}%
  \endgroup
}

\newcolumntype{L}[1]{>{\raggedright\arraybackslash}p{#1}}
\newcolumntype{Q}[1]{>{\centering\arraybackslash}p{#1}}

\definecolor{bluekeywords}{rgb}{0.13, 0.19, 0.7}
\definecolor{greencomments}{rgb}{0.1, 0.5, 0.2}
\definecolor{redstrings}{rgb}{0.8, 0.15, 0.1}
\definecolor{graynumbers}{rgb}{0.5, 0.5, 0.5}
\definecolor{subtlegray}{rgb}{0.98, 0.98, 0.98}
\usepackage{lstautogobble}
\usepackage{listings}

\usepackage{makecell} %

\usepackage{pifont}
\newcommand{\cmark}{\ding{52}}%
\newcommand{\xmark}{\ding{54}}%

\usepackage{hyperref}
\usepackage{balance}
\usepackage{flushend}

\def\BibTeX{{\rm B\kern-.05em{\sc i\kern-.025em b}\kern-.08em
    T\kern-.1667em\lower.7ex\hbox{E}\kern-.125emX}}
\begin{document}

\renewcommand\tabularxcolumn[1]{m{#1}} 

\newcolumntype{C}{>{\centering\arraybackslash}X}

\title{Bifrost: End-to-End Evaluation and Optimization of Reconfigurable DNN Accelerators}

\author{Axel Stjerngren, Perry Gibson, Jos\'e Cano \\
\emph{School of Computing Science, University of Glasgow, United Kingdom}
}

\maketitle

\begin{abstract}

Reconfigurable accelerators for deep neural networks (DNNs) promise to improve performance such as inference latency. STONNE is the first cycle-accurate simulator for reconfigurable DNN inference accelerators which allows for the exploration of accelerator designs and configuration space. However, preparing models for evaluation and exploring configuration space in STONNE is a manual developer-time-consuming process, which is a barrier for research.

This paper introduces \emph{Bifrost}, an end-to-end framework for the evaluation and optimization of reconfigurable DNN inference accelerators. \emph{Bifrost} operates as a frontend for STONNE and leverages the TVM deep learning compiler stack to parse models and automate offloading of accelerated computations. 
We discuss Bifrost's advantages over STONNE and other tools, and evaluate the MAERI and SIGMA architectures using Bifrost. 
Additionally, \emph{Bifrost} introduces a module leveraging AutoTVM to efficiently explore accelerator designs and dataflow mapping space to optimize performance.
This is demonstrated by tuning the MAERI architecture and generating efficient dataflow mappings for AlexNet, obtaining an average speedup of $50\times$ for the convolutional layers and $11\times$ for the fully connected layers.
Our code is available at \url{www.github.com/gicLAB/bifrost}.

\end{abstract}


\section{Introduction}

Deploying deep neural networks (DNNs), e.g. when targeting constrained devices, can be prohibitive due to steep computational requirements of state-of-the-art DNN models.
To address this issue, an across-stack approach is needed~\cite{iiswc_2018}, with algorithmic improvements giving better accuracy with fewer operations~\cite{hernandez2020} and novel compression techniques reducing model size further~\cite{blalock2020}. 
DNN inference accelerators can bring improvements for the hardware layer of the systems stack, with reconfigurable accelerators such as MAERI~\cite{kwon2018maeri} and Eyeriss v2~\cite{chen2019} promising improved performance by adjusting logic paths for a given DNN model architecture. 
However, finding optimal hardware configurations is still an active area of research~\cite{9072480}.
STONNE~\cite{stonne2021iiswc}, a cycle-accurate simulator for DNN accelerators with reconfigurable dataflow patterns, allows researchers to explore the design space of flexible accelerator architectures.
However, it currently requires significant manual effort to use, such as the requirement to rewrite the PyTorch model definition so it can be parsed by the system, as well as being limited to PyTorch support only. Additionally, the mappping tools to generate optimized dataflow configurations are not directly integrated in STONNE, such as mRNA for MAERI~\cite{zhao2019mrna}, and thus require further manual steps.

To address these usability issues of STONNE this paper proposes \emph{Bifrost}, a new tool that enables accessible end-to-end evaluation and optimization of reconfigurable DNN inference accelerators.
As well as automating many of the more tedious manual steps of using STONNE, \emph{Bifrost} also allows more DNN models to be run, adds a module for automatically generating optimized mappings for reconfigurable accelerators, as well as the ability to leverage existing mapping tools.

\emph{Bifrost} is built on STONNE and Apache TVM~\cite{chen2018tvm}, a state-of-the-art machine learning compiler framework that enables researchers to transparently execute any of the wide-range of DNN models compatible with TVM (from frameworks such as PyTorch~\cite{paszke2019pytorch}, TensorFlow~\cite{abadi2016tensorflow}, and ONNX~\cite{bai2019}) using STONNE. 
DNN layers not accelerated by the chosen hardware accelerator in STONNE are executed using an implementation from TVM, which allows end-to-end evaluation and easy verification of correctness.
Bifrost also extends the auto-tuning module of TVM, AutoTVM~\cite{chen2018c}, for design space and dataflow exploration, for example varying tile sizes to reduce clock cycle counts.
Additionally, Bifrost can integrate specialized mapping tools such as mRNA~\cite{zhao2019mrna} for MAERI, which may provide more optimal mappings in less time assuming that a specialized mapping tool is available for the target hardware architecture.
Note that Bifrost can be easily extended to support new accelerator architectures in STONNE including those with additional layer types.
The STONNE project is integrating power and area metrics, which Bifrost will support when they are available.

The main contributions of this paper include the following:

\begin{itemize}
    \item We motivate the need for Bifrost and its value for exploring reconfigurable DNN accelerators comparing it against a range of related DNN accelerator simulators.
    
    \item We describe in detail the features and implementation of Bifrost, such as how it integrates into TVM and the configuration options available.

    \item We describe how we automate many of the most tedious steps of the STONNE workflow, allowing models from more frameworks to be evaluated and opening the door to further compiler-hardware co-design exploration.

    \item We enable a new way to optimize mappings for reconfigurable STONNE accelerators by exposing tunable hardware parameters to AutoTVM. 
    In addition, we show how Bifrost can integrate specialized mapping tools such as mRNA, which can provide more efficient mappings. %
    
    \item We evaluate the layers of AlexNet~\cite{krizhevsky2012imagenet} using the SIGMA and MAERI architectures with varying levels of sparsity and approaches to mapping configuration respectively, to highlight the key functionality of Bifrost.

\end{itemize}

\section{Background}

The energy efficiency and performance of an DNN accelerator is determined by its \emph{dataflow}~\cite{samajdar2020,kwon2018maeri,chen2017}. 
Unlike server class GPUs and TPUs, edge accelerators devices do not have the hardware resources to process a DNN layer in a single step\footnote{Except for trivially small layers, which are not common in most production DNN models today.}. 
Instead the computation has to be divided up by grouping neurons into \emph{tiles} which defines how a group of neurons' inputs, weights, and intermediate outputs (\emph{psums}) are delivered and reused within the accelerator. 
This pattern is known as the \emph{dataflow} of the accelerator, which can vary among accelerator architectures.

The first generation of DNN accelerators have fixed dataflows tailored specifically for one type of workloads, e.g. systolic arrays (TPUs).
The next generation of DNN accelerators are reconfigurable, which means that aspects of their dataflow can be changed by software to increase efficiency (e.g., in terms of clock cycles, or energy consumption). A \emph{mapping} is a specific instance of a dataflow~\cite{krishna2020data} for reconfigurable accelerator architectures.  

Reconfigurable accelerators can be configured to map different dataflows and adjusting logic paths for a given DNN model architecture.
Being able to reconfigure the dataflow of the accelerator is especially useful in edge devices where the requirements to optimize inference time or performance per watt are more critical. 
MAERI~\cite{kwon2018maeri}, SIGMA~\cite{qin2020sigma}, MAGMA~\cite{nichols2019magmadnn}, and Eyeriss V2~\cite{9072480} are different examples of DNN accelerators with reconfigurable accelerator fabrics.
Reconfigurable accelerators are also more complex than fixed accelerators which results in a large mapping space.
Finding optimal hardware configurations for said accelerators is still an active area of research and the most common approach is to use analytical solutions.
This has been explored by Yu et al. for arbitrary accelerator designs~\cite{9072480}. 
Zhao et al. created a tool called mRNA to find optimal dataflow configurations for MAERI~\cite{zhao2019mrna}.

\subsection{STONNE}
\label{subsec:stonne}

STONNE~\cite{stonne2021iiswc} is the first cycle-accurate simulator for DNN accelerators with reconfigurable dataflow patterns that allows researchers to explore the design space of reconfigurable accelerator architectures.
As Krishna et al. point out, the \emph{mapping space} has to be separated from the \emph{architecture design space} when dealing with flexible DNN accelerators~\cite{krishna2020data}. 
For a given workload executed using a fixed DNN accelerator design, the performance and energy efficiency are solely determined by the physical features of the architecture (such as the number of processing elements). 
However, when a given workload is executed using a reconfigurable DNN accelerator the performance and energy efficiency will vary depending on the data flow as well the physical properties of the architecture.
A \emph{mapping} is a characterization of the scheduling and data orchestration of a reconfigurable DNN which determines the data flow. 

To date STONNE is able to simulate the reconfigurable accelerators MAERI, SIGMA, MAGMA, and a fixed systolic array (TPU).
Architectural simulators are common when developing and researching GPUs and CPUs, but for DNN accelerators STONNE is the first of its kind as it is able to efficiently simulate multiple fixed and reconfigurable accelerator designs. 
STONNE allows researchers to explore the performance and energy efficiency for different architecture design and mapping combinations.
However, preparing models for evaluation and exploring configuration space in STONNE is a manual developer-time-consuming process, which is a barrier for research.

\begin{figure}[t]
  \centering
  \includegraphics[width=0.95\linewidth]{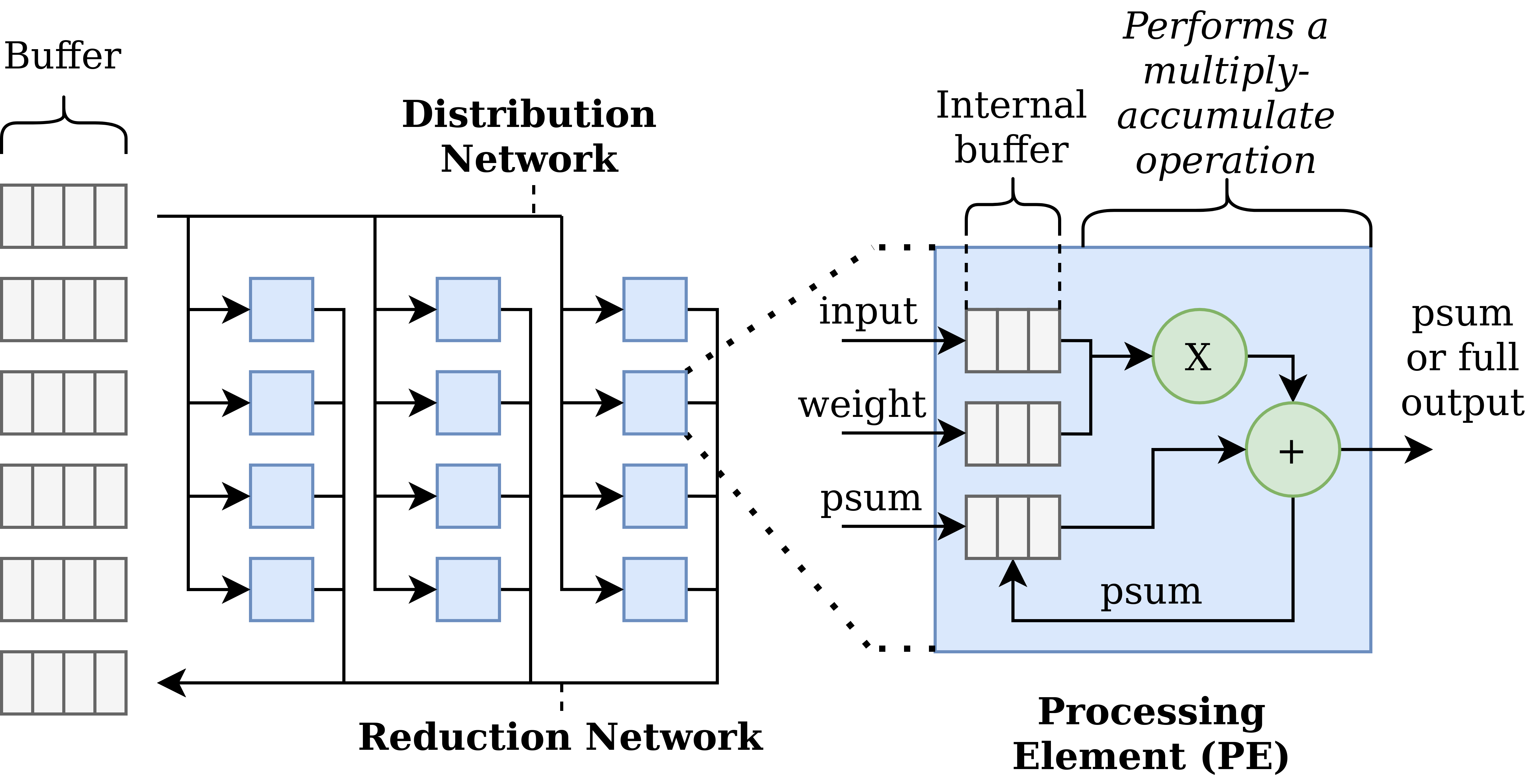}
  \caption{A simplified view of the main components of a typical reconfigurable DNN accelerator design~\cite{krishna2020data}.
  }
  \label{fig:dnnacellstonne}
\end{figure}

The DNN accelerator architectures simulated in STONNE all have the same basic components and these are illustrated in Figure~\ref{fig:dnnacellstonne}. 
The general structure of a DNN accelerator comprises of a spatial array of processing elements (PEs). 
Each PE contains a multiply-accumulate unit (MAC).
The PEs receive their inputs and weights from the distribution network and write outputs back to the buffer using the reduction network. 
A MAC operation involves the computation of the product of two numbers $b$ and $c$ and adding the product to an accumulator $a$, that is: $a\leftarrow a+(b\times c)$. 
The intermediate outputs which are computed through a MAC operation are called partial sums or \emph{psums}.

\begin{figure}[t]
  \centering
    \includegraphics[width=\linewidth]{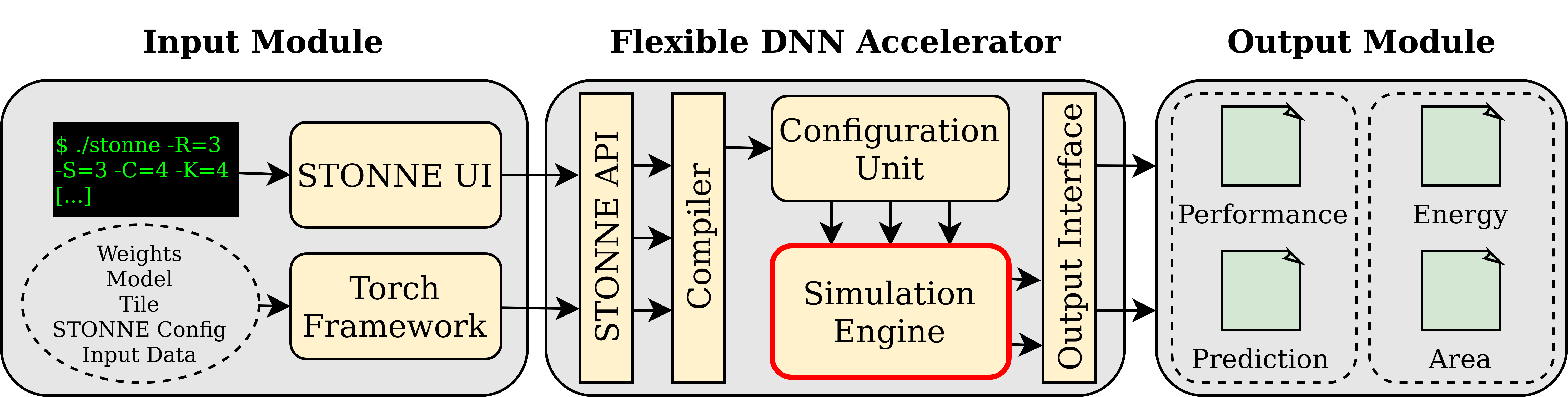}
  \caption{Overview of STONNE, adapted from~\cite{stonne2021iiswc}.} 
  \label{fig:stonne}
\end{figure}

\subsection{TVM}
\label{subsec:tvm}

\begin{figure}[t]
  \centering
  \includegraphics[width=0.95\linewidth]{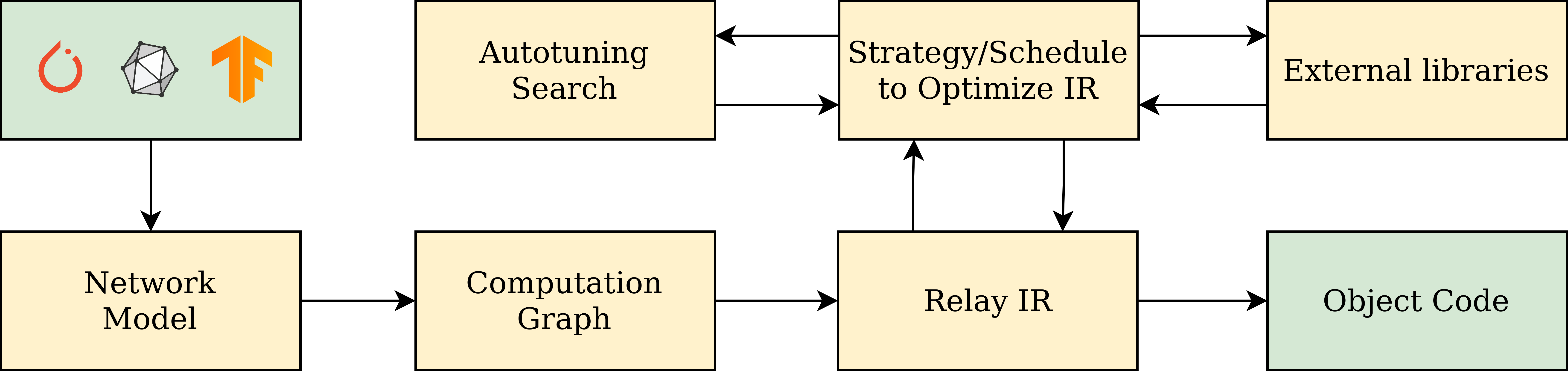}
  \caption{Simplified overview of TVM.
  Models can be loaded for a variety of sources, and compiled to a binary.
  Optionally, auto-tuning can occur, or external libraries can be leveraged.
  } 
  \label{fig:tvm_simple}
\end{figure}

Apache TVM~\cite{chen2018tvm} is an end-to-end machine learning compiler framework for CPUs, GPUs, and accelerators. 
TVM calls these different runtime backends \emph{targets}. 
TVM is compatible with models from a variety of deep learning frameworks such as PyTorch~\cite{paszke2019pytorch}, TensorFlow~\cite{abadi2016tensorflow}, TensorFlow Lite, and ONNX~\cite{bai2019}. 
In general, deep learning frameworks use computational graphs as their intermediate representation, which are directed acyclic graphs (DAGs) representing each step in the computation process. 
TVM can parse models from deep learning frameworks and translate it to its own intermediary representation, Relay IR~\cite{DBLP:journals/corr/abs-1810-00952}
A simplified overview of TVM is shown in Figure~\ref{fig:tvm_simple}.

Each node in the Relay IR requires a corresponding operator (called compute and schedule functions) to execute the computation in the node.
These operators are stored in the TVM Operator Inventory (TOPI) and are specialized for each target. Operators can also be provided by \emph{external libraries} where TVM will transparently offload computations to the library.
Often there are several different algorithms and implementations available for any given operator, and TVM uses a Relay Operator Strategy function to select which operator to use. 
For example, different memory layouts for convolutions require different operators.
The strategy then calls operators from the TOPI or from an external library. The operators from TOPI are implemented in TVM's internal Tensor Expression Language.
These expressions are then used to select schedule primitives which are used to generate the low level code. 
To further optimize the schedule, parameters such as tile size, loop ordering, and re-ordering can be explored using the AutoTVM auto-tuning module~\cite{chen2018tvm}. 
AutoTVM automatically optimizes the schedule using the latency of the computation and developers are able to declare tunable parameters called \emph{tuning knobs} in the schedule space.

\section{Motivation}

As shown in Figure~\ref{fig:stonne}, STONNE can provide information on the performance of the accelerator (cycle count), and in future the energy and area used. 
However, using STONNE for research is a time consuming process with many manual steps. 
Figure~\ref{fig:flowchart_stonne} demonstrates the typical workflow using STONNE, which we discuss in more detail in Section~\ref{subsec:stonne_workflow}.
Section~\ref{subsec:other_systems} highlights some valuable features that a simulator for reconfigurable DNN inference accelerators should have, and provides a table comparing some existing tools and systems.
Section~\ref{subsec:other_systems} highlights some valuable features that a simulator for reconfigurable DNN inference accelerators should have, discusses previous works/systems, and summarizes the differences according to the features in a table.

\subsection{STONNE workflow}
\label{subsec:stonne_workflow}

\begin{figure}[t]
  \centering
  \includegraphics[width=0.95\linewidth]{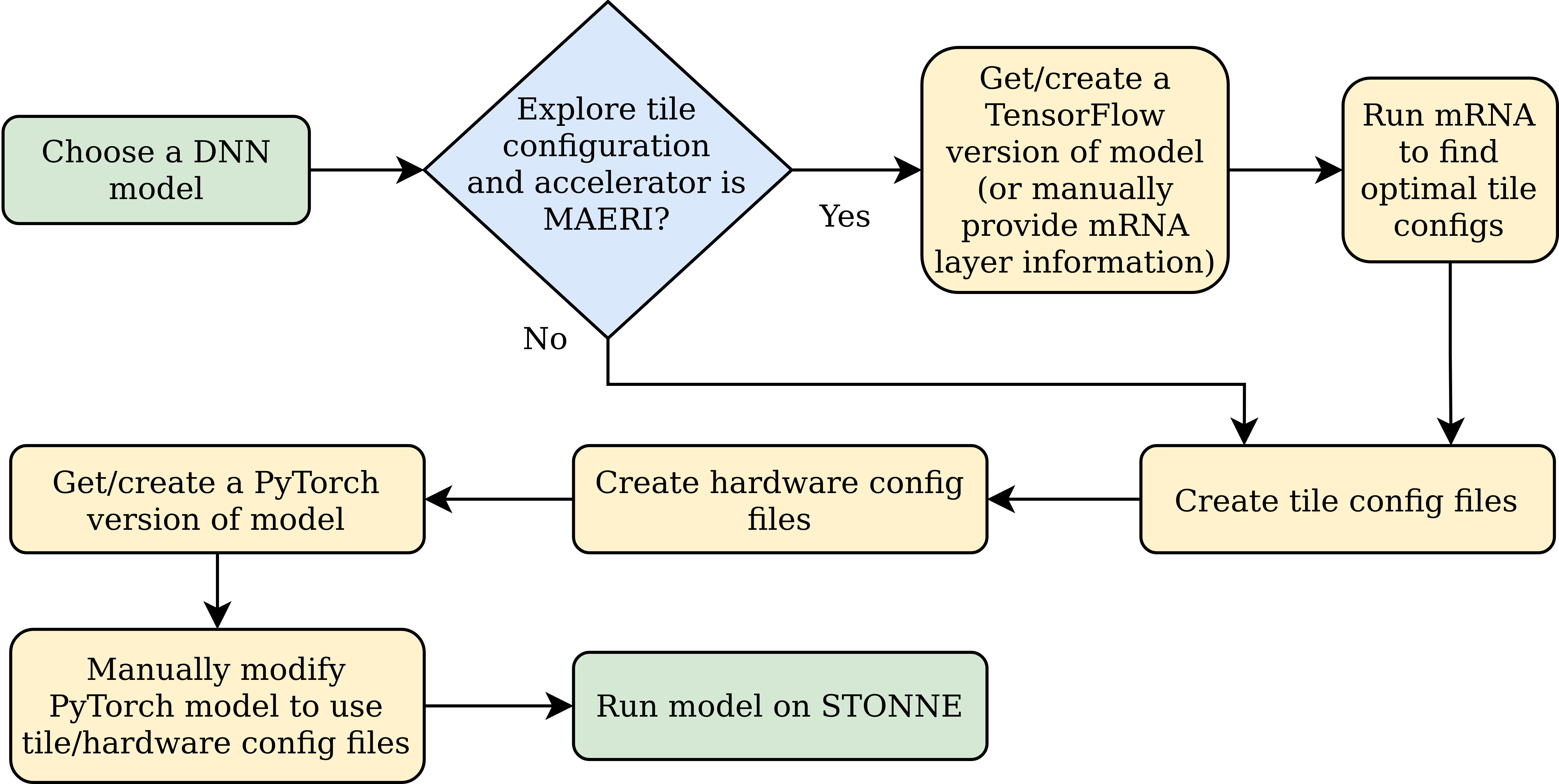}
  \caption{
Flowchart of the typical operation of STONNE. Note the manual steps, and the requirement for both PyTorch and TensorFlow versions of the model if exploring the mapping.
  \label{fig:flowchart_stonne}}
\end{figure}

The different steps shown in Figure~\ref{fig:flowchart_stonne} are explained in detail:

\begin{enumerate}
    \item \textbf{Choose a DNN model:} The first step is to choose a model to explore.
    Currently, accelerator architectures in STONNE support models containing 2D convolutional and/or fully connected layers.

    \item \textbf{Explore mapping configuration:} While STONNE is able to simulate reconfigurable accelerators, the onus is still on the user to find an optimal dataflow mapping.
    Some accelerator designs have tools available to find optimal mappings, such as mRNA~\cite{zhao2019mrna} for the MAERI accelerator architecture.
    However these external tools are not necessarily compatible with PyTorch, the only deep learning framework which is fully compatible with STONNE.
    For example, mRNA only supports TensorFlow models which are not supported by STONNE, meaning two versions of the model are needed to generate a mapping.
    Thus, users must either convert the models (which is not necessarily an easy process, even with interchange formats like ONNX being available), or find/create native definitions for the model in both frameworks.
    
    \item \textbf{Create tile (mapping) config files:} Each mapping requires a corresponding test file to be manually created. 
    
    \item \textbf{Create hardware config files:} The user defines the hardware resources they want to give to their chosen accelerator (e.g., number of PEs).
    Using STONNE to explore the performance of different hardware parameter configurations is possible, but requires the user to create a configuration file for every hardware variation and then to run STONNE manually with each hardware configuration.
    
    \item \textbf{Get/create a PyTorch version of model}: STONNE currently only supports PyTorch, with some limited (deprecated) support for Caffe.
    
    \item \textbf{Manually modify PyTorch model to use mapping and hardware configuration files:} STONNE is prepackaged with a forked version of PyTorch which contains extra operators to run a layer using STONNE.
    
    \item \textbf{Run a model on STONNE}: with the mapping and hardware config files, the modified PyTorch model can be executed on STONNE, which will provide metrics on the simulated inference.
\end{enumerate}

Observing this workflow, we conclude that research using STONNE is possible but it requires significant manual effort. 
STONNE itself may be a valuable platform for developing new reconfigurable deep learning inference accelerators, however there are a number of ways in which its workflow could be improved. 
The main limitations of STONNE is the lack of support for running models from deep learning frameworks beyond PyTorch and the need for external tools to find optimal hardware configurations.
In addition, for MAERI's mapping tool mRNA we require a TensorFlow version of the model.
Integrating STONNE and a deep learning compiler such as TVM into a unified framework would open the door to further research into DNN model/hardware co-design.
This could bring performance improvements and reduce mapping costs, as well as enable more complex reconfigurable accelerator designs.
We address these problems with our new tool \emph{Bifrost}, which achieves a significantly improved developer workflow by integrating STONNE together with Apache TVM.

\subsection{Comparison against related works/systems}
\label{subsec:other_systems}

There is a variety of DNN accelerator simulators available, some with support for reconfigurable architectures, others with fixed accelerators.
To compare them and highlight the contribution of \emph{Bifrost} as a tool, we identify 6 valuable features for reconfigurable DNN hardware accelerator simulators and compare several other works against these features:

\begin{enumerate}[label=\roman*)]
    \item \emph{Model support}: the ease of using DNN models from a wide range of DNN frameworks (e.g., PyTorch, TensorFlow, ONNX, etc).
    
    \item \emph{Easy mapping exploration}: if the system provides tuning support for reconfigurable accelerator architectures.
    
    \item \emph{Multiple accelerator architectures available}: if more than one DNN accelerator architecture is available in the system.
    
    \item \emph{Sparsity support}: if sparse inference is available for DNN models, i.e. reducing inference costs by skipping MAC operations involving zeros.
    
    \item \emph{Mainstream DNN framework integration}: if the system is well integrated within a mainstream DNN framework (such as PyTorch, TensorFlow, TVM, etc), which brings the advantages of a large community, frequent updates, and troubleshooting.
    
    \item \emph{Cycle-accurate Simulation Available}: if the system provides cycle-accurate simulation.
\end{enumerate}

Table~\ref{tab:taxonomy2} compares various related tools and systems to \emph{Bifrost}, using the above features.
Next, we discuss the details of each of the tools and systems from the table.

\textbf{SMAUG}~\cite{xi2020} provides a full simulation-based system that uses gem5-Aladdin~\cite{shao2016} to perform full system simulation of the host system, the off-chip memory accesses, and the accelerator itself.
SMAUG does not integrate with existing DNN frameworks, instead models must be redefined using SMAUG's Python API.
It can also support accelerators with sparse inference.

\textbf{SCALE-Sim}~\cite{samajdar2019} is a cycle-accurate simulator framework which provides configurable systolic array designs, with users defining a config file describing their chosen architecture.
It does not integrate with existing DNN frameworks, nor provide end-to-end model evaluation.
Instead the user must define their network configuration as a DNN topology file to be parsed by the tool.

\textbf{SECDA}~\cite{Haris2021SBACPAD} is a DNN accelerator design methodology leveraging SystemC.
It uses transaction-level simulation (rather than cycle accurate), synthesizing designs on real hardware to get more accurate system performance metrics.
The authors provide 2 case studies integrated into the TFLite DNN inference framework, neither of which provide sparse inference.

\textbf{VTA}~\cite{moreau2018vta} is a DNN accelerator architecture officially integrated into TVM.
This integration provides the advantages of TVM, such as support for models from most DNN frameworks.
The accelerator design uses an ISA, which means that the compiler generates instructions for a given layer to run on the accelerator.
This means that the compiler can generate more or less optimal instructions, however this does not fit the definition of a dataflow mapping.

\textbf{STONNE}~\cite{stonne2021iiswc} is a DNN accelerator tool designed for use with reconfigurable DNN accelerator designs such as MAERI.
To date, it supports 3 reconfigurable accelerator architectures (MAERI, SIGMA, and MAGMA) and 1 fixed accelerator architecture (a TPU), with one of the architectures (SIGMA) supporting sparse inference.
Exploring mapping space and running models is not a straightforward process, as discussed in Section~\ref{subsec:stonne}.

\textbf{Bifrost} combines the advantages of STONNE, within the TVM framework.
It extends the AutoTVM auto-tuning module to provide automatic accelerator configuration search, whereas STONNE must rely on external tools (which Bifrost also integrates).
Bifrost's value is for users who want to explore the potential of reconfigurable accelerators (as provided by STONNE), however want increased productivity by automating many of the more tedious steps.
Since STONNE is an open source tool, Bifrost has the potential to improve the ease of testing and improving new accelerator designs in STONNE.
This is due to its AutoTVM module, which can search for optimized mappings even when no specialized mapping tool is available, although these specialized mapping tools can be integrated.
Note that in the future, the integration with TVM opens the door to further compiler-hardware co-design exploration, something that we leave for future work.

\begin{table}[t]
\vspace{1.5mm}

    \caption{
Comparison of Bifrost against other tools and systems for DNN accelerator evaluation.}\label{tab:taxonomy2}
  \centering
  {
\begin{tabular}{|l||r|r|r|r|r|r|}
    \hline
    \backslashbox{\textbf{Feature}\quad\quad}{\rotatebox[origin=c]{90}{\hspace{0.5mm}\textbf{System}\quad\quad\quad}} &	\rotatebox[origin=c]{90}{\textbf{SMAUG}~\cite{xi2020}}	&	\rotatebox[origin=c]{90}{\hspace{0.5mm}\textbf{SCALE-Sim}~\cite{samajdar2019}}	&	\rotatebox[origin=c]{90}{\textbf{SECDA}\cite{Haris2021SBACPAD}}	&	\rotatebox[origin=c]{90}{\textbf{VTA}~\cite{moreau2018vta}}	&	\rotatebox[origin=c]{90}{\textbf{STONNE}~\cite{stonne2021iiswc}}	&	\rotatebox[origin=c]{90}{\textbf{Bifrost}}	\\
    \hline
    Model support	&	\xmark	&	\xmark	&	\xmark	&	\cmark	&	\xmark	&	\cmark	\\
    Easy mapping exploration	&	\xmark	&	\xmark	&	\xmark	&	\xmark	&	\xmark	&	\cmark	\\
    Multiple accelerators	&	\cmark	&	\cmark	&	\cmark	&	\xmark	&	\cmark	&	\cmark	\\
    Sparsity support	&	\cmark	&	\xmark	&	\xmark	&	\xmark	&	\cmark	&	\cmark	\\
    DNN framework integration	&	\xmark	&	\xmark	&	\cmark	&	\cmark	&	\xmark	&	\cmark	\\
    Cycle-accurate simulation	&	\cmark	&	\cmark	&	\xmark	&	\xmark	&	\cmark	&	\cmark	\\
    \hline
\end{tabular}}

\end{table}

\section{Bifrost Overview}

\begin{figure}[t]
  \centering
    \includegraphics[width=0.85\linewidth]{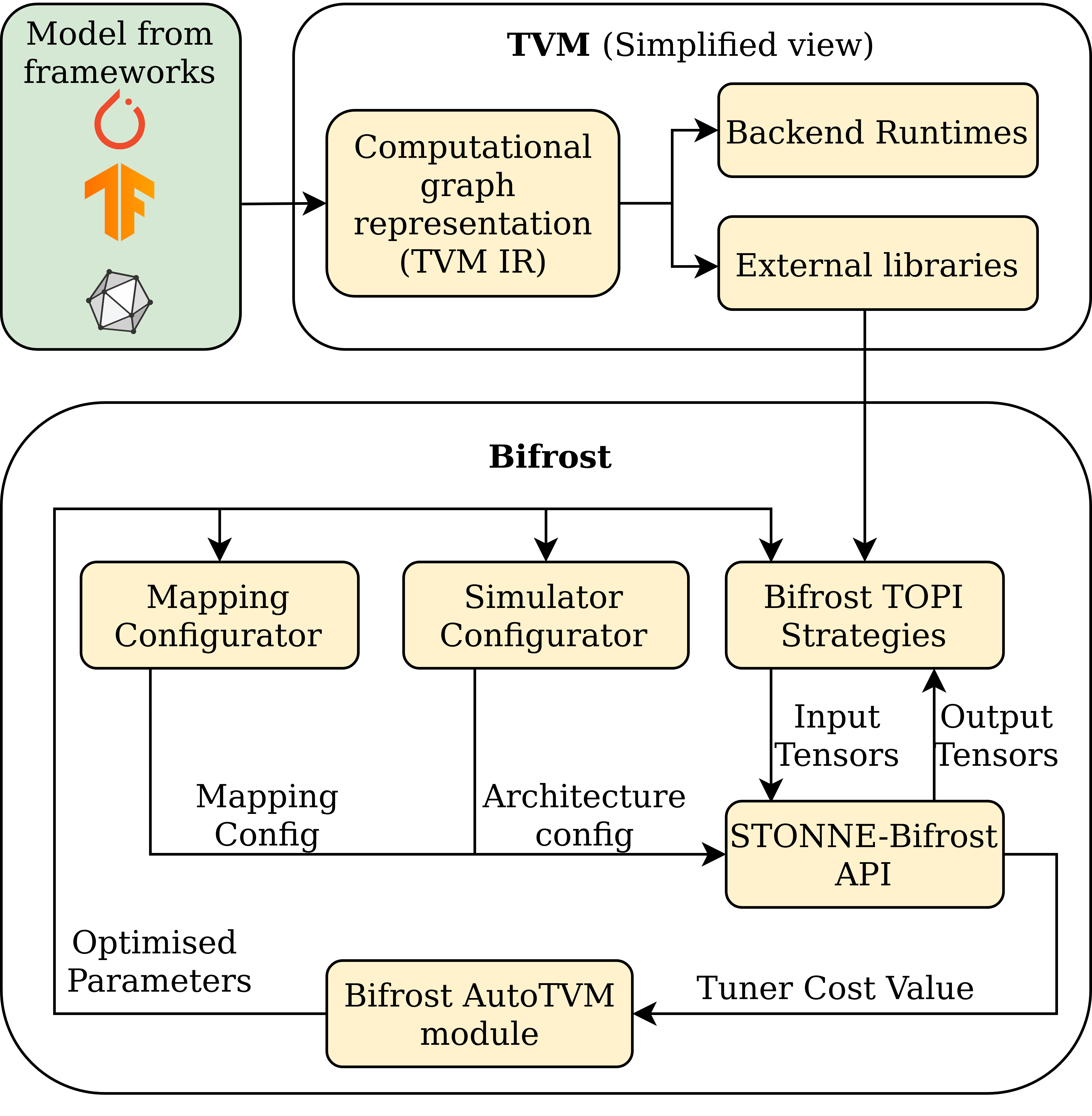}
  \caption{High-level overview of \emph{Bifrost} design.}
  \label{fig:overview}
\end{figure}

\emph{Bifrost} is our proposed solution to connect the STONNE hardware accelerator simulator to the TVM compiler framework.
Our goal is to make the STONNE tool for evaluating simulated hardware accelerators for DNNs as simple to use as possible, while enabling additional functionality such as an accelerator architecture agnostic mapping generator leveraging TVM. 
Figure~\ref{fig:overview} gives a high-level overview of \emph{Bifrost}.
First, the user provides a DNN model from any deep learning framework supported by TVM (such as PyTorch, TensorFlow, MXNet, Keras, TFLite, Caffe2, or ONNX). 
This improves significantly on STONNE, which only has support for PyTorch models. 
Then TVM parses the model, translating the computation graph into its own intermediate representation, applies some graph-level optimizations (e.g., fusion of batch normalization layers), and chooses the backend for each operation using its \emph{TOPI} library.
For operations supported by \emph{Bifrost}, which are currently 2D convolutional layers (conv2d) and dense (fully connected) layers, TVM TOPI offloads operations with calls to the STONNE-Bifrost API (discussed in Section~\ref{sec:bifrost-api}) as an external library, which sends all relevant layer information to STONNE while TVM uses its own code generation for non-accelerated layers.
This limitation in operations is inherited by the accelerators designs currently available in STONNE, and it is straightforward to add support to new operations when a new STONNE accelerator requires them.

Since STONNE can simulate a wide range of accelerator architectures, the user is able to specify the architecture and mapping used for running the DNN model, although these steps require more manual configuration which Bifrost automates. 
STONNE can simulate a variety of inference accelerators, and to-date \emph{Bifrost} supports MAERI~\cite{kwon2018maeri}, SIGMA~\cite{qin2020sigma}, and a systolic array (i.e., a TPU~\cite{jouppi2017}), with more accelerators to be added as the STONNE community develops them. 
For reconfigurable accelerators (such as MAERI) \emph{Bifrost} implements a mapping tool to find configurations for the hardware given a DNN model. 
This mapping tool leverages AutoTVM to find a mapping for any accelerator which exposes tunable parameters.
In addition, \emph{Bifrost} also supports and integrates existing mapping tools such as mRNA for MAERI.

The key components of \emph{Bifrost} (Figure~\ref{fig:overview}) are listed below:

\begin{enumerate}
    \item \textbf{STONNE-Bifrost API} A C++ library which processes layer information from the custom TOPI strategies and uses it to configure STONNE.
    
    \item \textbf{Bifrost TOPI strategies} Act as the bridge between TVM and STONNE by passing all relevant layer information to the \emph{STONNE-Bifrost API}.
    
    \item \textbf{Simulator Configurator} Allows users to programmatically specify the simulated architecture on STONNE and ensures that only valid hardware configurations for simulation are specified. Hardware configurations can be tuned using the AutoTVM module.
    
    \item \textbf{Mapping Configurator} Specifies the dataflow of each layer. 
    Mappings can be provided manually, or a default configuration can be automatically generated, or the mapping can be tuned with the AutoTVM module, or another specialized tool such as mRNA.
    
    \item \textbf{Bifrost AutoTVM Module} Explores hardware configuration space for a given DNN model, adjusting the hardware configuration parameters exposed to it via the API.
\end{enumerate}

\begin{figure}[t]
  \centering
\includegraphics[width=0.9\linewidth]{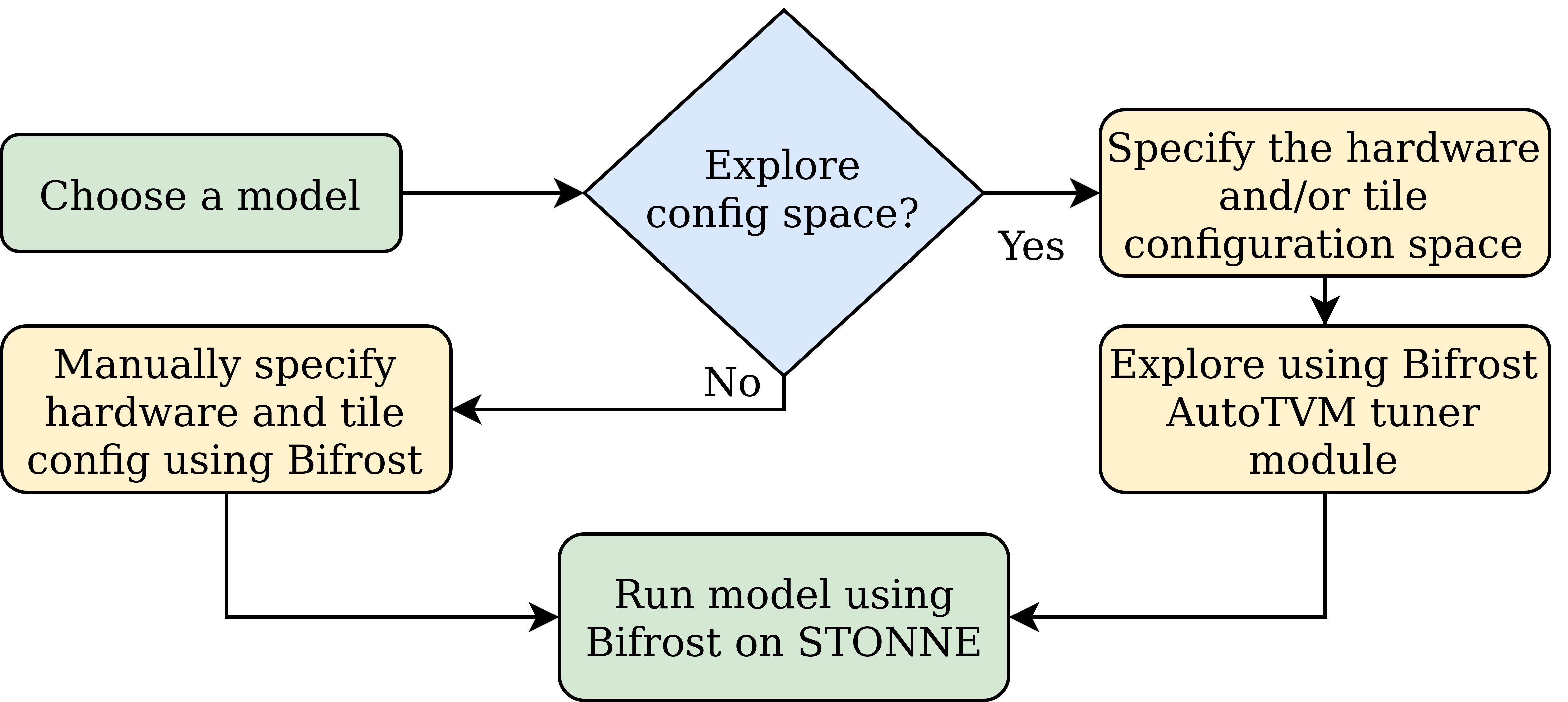}

  \caption{Flowchart of the operation of STONNE using \emph{Bifrost}.  
  \label{fig:flowchart}}
\end{figure}

Figure~\ref{fig:flowchart} demonstrates the \emph{Bifrost} workflow for hardware design space exploration. Executing a module in \emph{Bifrost} is demonstrated in Listing~\ref{lst:bifrostmodel}.
Note how the full DNN model is transparently executed without any modification, in comparison to STONNE's standard workflow in Figure~\ref{fig:flowchart_stonne}.

In principle, Bifrost can also work with physical hardware as long as it exposes the same API as the STONNE accelerator version of the same hardware.
However, since most of these reconfigurable accelerators are not yet available for evaluation in real hardware and do not have mature toolchains (e.g., drivers), for now we cannot evaluate them.
Reconfigurable DNN accelerators are a burgeoning area of hardware design, with STONNE positioning itself as a tool to enable the design and implementation of new designs.
Bifrost complements this by automating more tedious steps of evaluation and providing a hardware configuration exploration tool which can bring value when no specialized tool (like mRNA) is available.

\begin{lstlisting}[language=python, float, caption={Running an arbitrary PyTorch model using Bifrost.}, label=lst:bifrostmodel]
import bifrost
from bifrost.simulator import architecture

// Set the amount of mulitpliers
architecture.ms_size = 128
architecture.create_config_file()

from bifrost.runner import run_torch_stonne
from neuralnet import model, input_batch

out = run_torch_stonne(model, input_batch)
\end{lstlisting}

\section{The STONNE-\emph{Bifrost} API}
\label{sec:bifrost-api}

The STONNE-\emph{Bifrost} API is where layer information such as height, width, strides, and padding are processed by TVM together with the architecture and dataflow mapping. 
This information is then used to execute the layer in STONNE and the output is passed back to the TVM Python frontend.
The API contains a set of \emph{packed functions}, which is the unified function type of TVM.
These type erased functions are made available in TVM's global function registry. When registered, a given function is automatically exposed to the TVM Python frontend when the API is loaded using Python \emph{ctypes} (a foreign function library). 

For example, the function to execute \texttt{NCHW} convolutions using STONNE is registered as \texttt{tvm.contrib.stonne.conv2d.nchw}. When the TVM frontend is executing a model, \texttt{tvm.contrib.stonne.conv2d.nchw} can be called to execute a convolutional layer using STONNE. The execution workflow for all functions in the API follow the same general pattern:

\begin{enumerate}
    \item Parse layer information.
    \item Transform layer information and input data into a format compatible with STONNE.
    \item Create a new instance of STONNE.
    \item Configure STONNE with the new architecture and dataflow mapping.
    \item Load the layer into STONNE and run.
    \item Transform output into a format compatible with TVM.
    \item Record the simulated cycle count and/or partial sums. 
\end{enumerate}

\emph{Bifrost} currently supports 2D convolutional and fully connected layers, the two main operations supported by STONNE.
Given that these two operations have been chosen as they are very computationally expensive layers and are commonly used in many DNNs, it makes sense that they are the target for hardware accelerators.
For example, when executing AlexNet~\cite{krizhevsky2012imagenet} (a popular convolutional neural network) on a GPU 95\% of the time is spent on the convolutional and fully connected layers~\cite{jia2014learning}, the other layers such as the pooling and activation functions only account for 5\% of the execution time.

\subsection{Fully Connected Layers (Dense)}

Fully connected layers are divided into two steps in TVM's computational graph, first a dense operator which applies a linear transformation followed by an optional non-linear activation function. 
Only the dense operator is executed on \emph{Bifrost} while the non-linear activation function is handled by the code generated by TVM for the CPU.
MAERI, SIGMA, and the TPU all implement the dense operator using a general matrix multiplication (GEMM).

In hardware architectures simulated using STONNE the execution will depend on the type of architecture and the (tile) mapping. 
For example, in MAERI architectures the tile pattern has to be provided as a parameter; in SIGMA architectures the memory controller automatically tiles the matrix depending on the level of sparsity~\cite{qin2020sigma}; and since the TPU has a fixed dataflow architecture, the tiling can not be changed.

\subsection{Convolutional layers (Conv2d)}

\begin{figure}[t]
  \centering
    \includegraphics[width=0.9\linewidth]{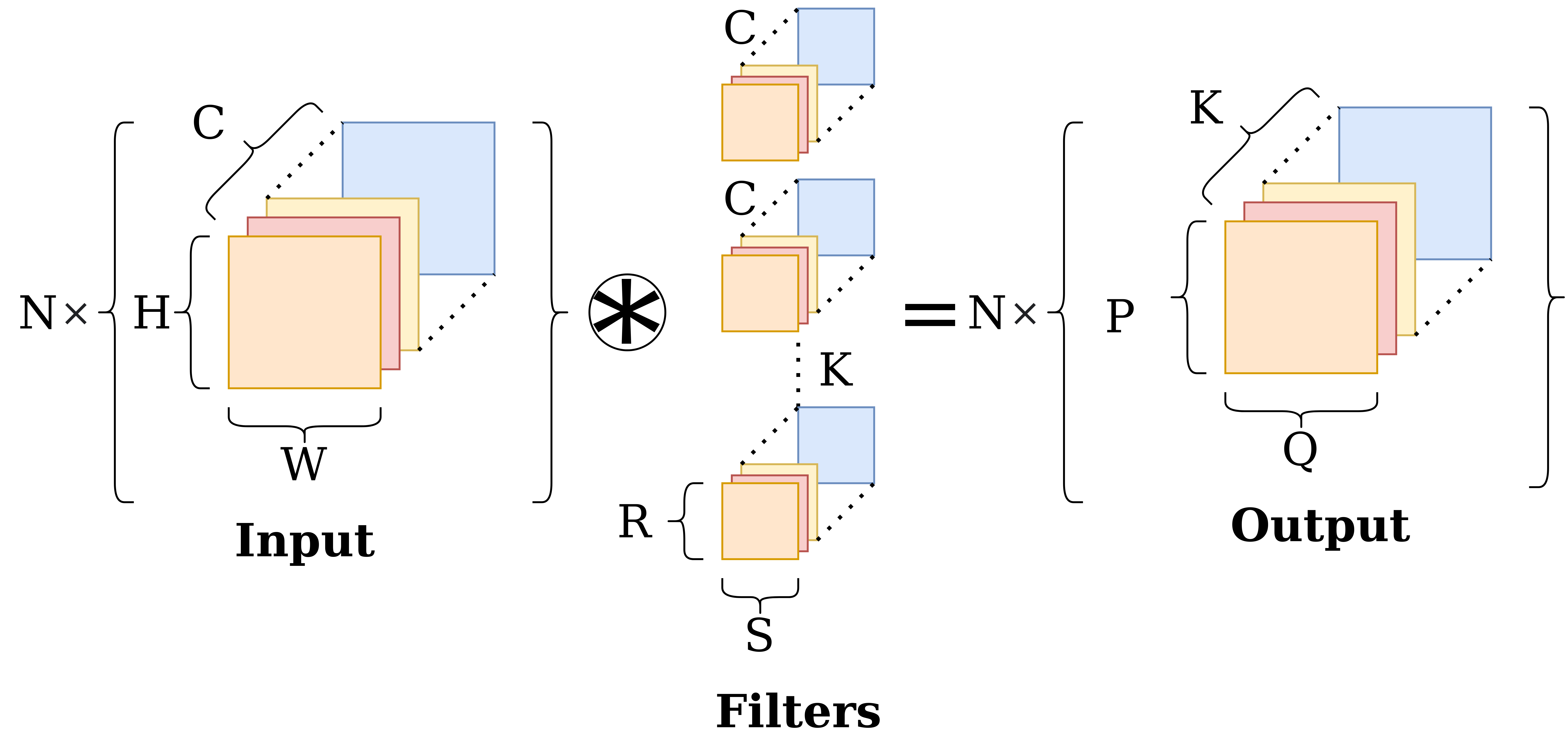}
  \caption{An \texttt{NCHW} convolution. Each channel is stored contiguously in memory and the kernel is stored in the \texttt{RSCK} format~\cite{nvidia}.}
  \label{fig:nchw}
\end{figure}

\begin{figure}[t]
  \centering
    \includegraphics[width=0.9\linewidth]{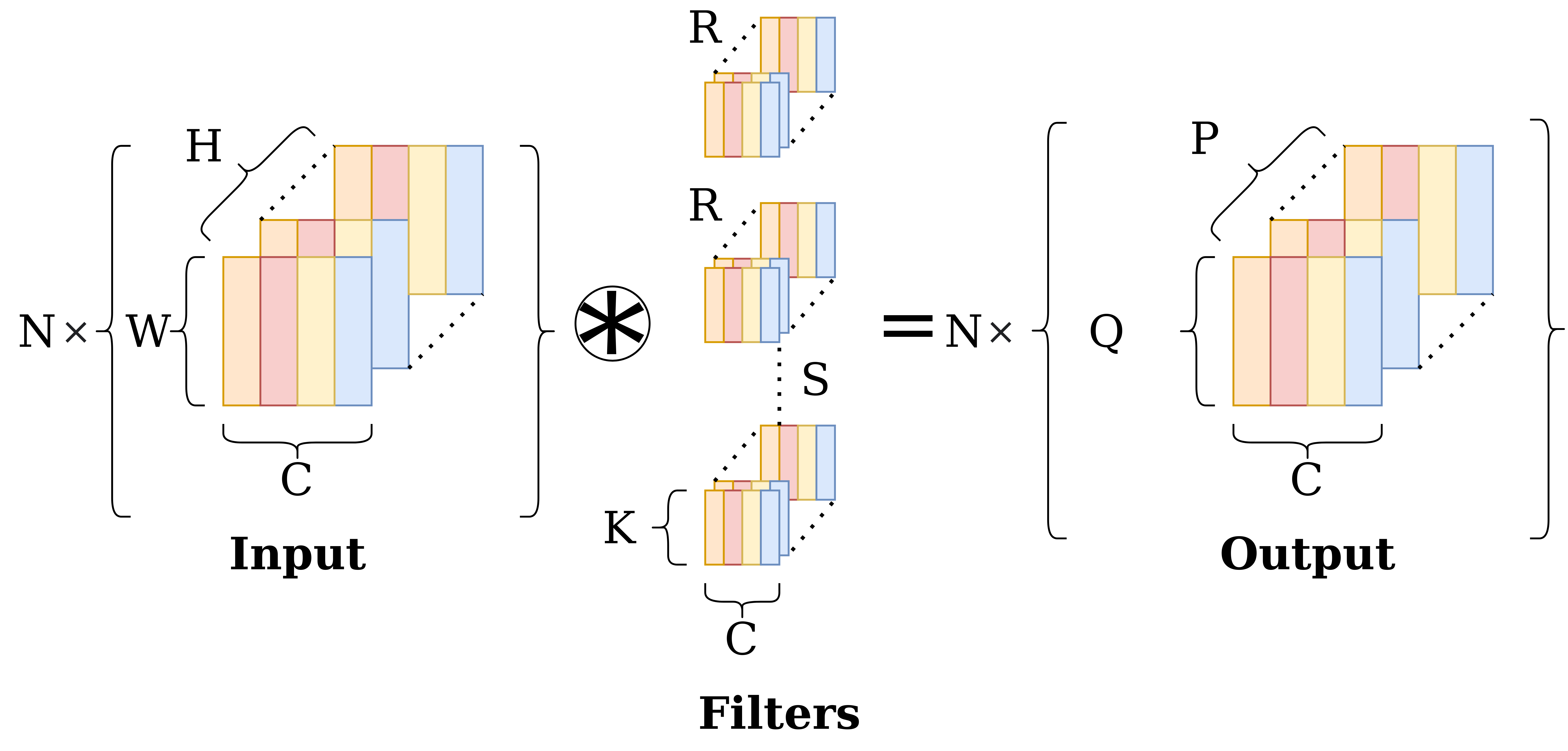}

  \caption{An \texttt{NHWC} convolution. The height and width are interleaved with the channels in memory and the kernel is stored in the \texttt{KCRS} format. }
  \label{fig:nhwc}
\end{figure}

The parameters that govern convolutions are listed in Table~\ref{tab:taxonomy}.
\emph{Bifrost} supports \texttt{NCHW} and \texttt{NHWC} 2D convolutions with \texttt{KCRS} and \texttt{RSCK} kernel layouts respectively.

\begin{table}[h]
\vspace{1.5mm}
    \caption{Standard parameters defining convolutional layers (Nvidia taxonomy)~\cite{nvidia}.}
    \label{tab:taxonomy}
    \centering
    \begin{tabular}{|c|c|}
    \hline
    \textbf{Parameter} & \textbf{Description}                \\
    \hline
    N & Batch size (STONNE only supports N=1) \\\hline
    R & Number of filter rows \\\hline
    S & Number of filter columns \\\hline
    C & Number of input channels \\\hline
    K & Number of output channels \\\hline
    G & Number of groups \\\hline
    H & Number of input rows \\\hline
    W & Number of input columns \\\hline
    P &  Number of output rows \\\hline
    Q &  Number of output columns \\\hline
    PadH & Height of zero-padding \\\hline
    PadW & Width of zero-padding \\
    \hline     
    
    \end{tabular}
\end{table}

An input tensor for a conv2d layer always consists of the same components: 
a number of batches ($N$), a number of channels $C$, height ($H$), and width ($W$).
However, these tensors can be stored in memory in a number of different ways, and deep learning frameworks have adopted different default layouts.
\texttt{NCHW} and \texttt{NHWC} are the data layouts used by default in PyTorch and TensorFlow respectively.
For tensors ordered using the \texttt{NCHW} layout each channel is stored contiguously in memory, while in the \texttt{NHWC} layout the height and width are interleaved with the channels.

Depending on the algorithmic primitive used, each input data layout format has a complementary kernel data layout format.
Changing either requires adjusting the algorithmic primitive used, and there are common data-layout/kernel-layout pairs used by deep learning libraries. 
A kernel tensor consists of a number of input ($C$) and output ($K$) channels, height ($R$), and width ($S$).
For an \texttt{NCHW} input the kernel is typically stored as \texttt{KCRS} while for \texttt{NHWC} inputs the kernel is typically stored as \texttt{RSCK}. 
Figure~\ref{fig:nchw} illustrates how a convolutional layer is executed with the \texttt{NCHW} layout, and Figure~\ref{fig:nhwc} illustrates how the same layer would be executed with the \texttt{NHWC} layout. 
TVM has support for both common layouts, and internally can use other layouts for more optimized inference such as spatial pack convolution~\cite{zheng2018a}.
Thus, the STONNE-\emph{Bifrost} API supports both formats and implements these through the \texttt{tvm.contrib.stonne.conv2d.nchw} and \texttt{tvm.contrib.stonne.conv2d.nhwc} functions.

\subsubsection{MAERI Convolutions}

The MAERI architecture on STONNE only supports \texttt{NHWC} convolutions with \texttt{RSCK} kernel layouts. 
If the input dimensions are \texttt{NHWC} the layer can be executed with minimal change to the data provided by TVM, as the input tensor only requires some padding to be added for STONNE compatibility. 
When the input dimensions are of the form \texttt{NCHW} and \texttt{KCRS}, the dimensions have to be transposed to be compatible with MAERI. 
This conversion is implemented in C++ and executed in the CPU, therefore the performance penalty of the conversion is not counted in the total cycle count for execution on STONNE.
The execution path for \texttt{NCHW} convolutions is as follows:

\begin{enumerate}[label=\roman*]
    \item The \texttt{NCHW} input is transposed to \texttt{NHCW}.
    
    \item The \texttt{KCRS} kernel is transposed to \texttt{RSCK}.
    
    \item A new instance of STONNE is created and configured with the chosen architecture and dataflow mapping. The \texttt{NHWC} and \texttt{RSCK} inputs are then fed into STONNE.
    
    \item The \texttt{NPQK} output is transformed to \texttt{NKPQ}.
\end{enumerate}

\subsubsection{SIGMA Convolutions}

The SIGMA architecture does not support convolutional layers.
SIGMA is a sparse accelerator architecture which only supports GEMM~\cite{qin2020sigma}.
However, it is possible to effectively convert the convolutions to a GEMM operation using an algorithmic primitive commonly known as GEMM convolution.
The input and weight tensors are converted from four dimensional tensors to 2D matrices. 
\texttt{NCHW} input tensors with \texttt{KCRS} kernels are multiplied together as $\mathit{weight} \times \mathit{data}$ while \texttt{NHWC} input tensors with \texttt{RSCK} kernels are multiplied together in the reverse $\mathit{data} \times \mathit{weight}$ order.

\subsubsection{TPU Convolutions} 

Like SIGMA, the TPU does not support convolutional layers directly.
Convolutional layers are instead executed using a GEMM operation. 
The TPU only supports \emph{NCHW} convolutions and the execution steps for a conv2d workload is as follows: 
\begin{enumerate}[label=\roman*]
    \item Transform the input and weight tensors into 2D matrices.
    \item Perform a GEMM operation and multiply the data and the weight matrices where the order of the operation depends on the input dimensions.
    \item Transform the 2D output data into the required 4D tensor.
\end{enumerate}

\section{Bifrost Hardware Configuration}

As illustrated in Figure~\ref{fig:stonne}, STONNE's \emph{configuration unit} uses configuration files to determine how the simulator should be configured. 
In \emph{Bifrost} this is handled by the \textbf{simulator configurator} module. 
All the hardware configuration options in STONNE and their possible values, which are currently available for use in \emph{Bifrost}, are listed in Table~\ref{tab:cfg}. 
Note that these options are supported by STONNE but the rules are \emph{enforced} by \emph{Bifrost}. 
By enforcing these rules, \emph{Bifrost} eliminates undefined behavior from occurring in STONNE by preventing developers from providing invalid hardware configurations. Note that not all configuration options are available for all accelerator architectures, however the simulator configurator will reject any invalid configurations. 
Below is a brief description of each hardware parameter and their associated restrictions:

\begin{enumerate}
    \item \textbf{controller\_type} is the architecture type such as MAERI, SIGMA, and TPU.
    
    \item \textbf{ms\_network\_type}. MAERI and SIGMA must use the \texttt{LINEAR} option while the TPU must use the \texttt{OS\_MESH} option, which means PEs are organized as a grid sending and receiving data using a weight-stationary dataflow.
    
    \item \textbf{ms\_size} is the number of multipliers (PEs) in the architecture.
    Each multiplier performs a MAC operation. More multipliers result in higher parallelism and performance. This parameter is used when the ms\_network\_type has been set to the \texttt{LINEAR} option.
    
    \item \textbf{ms\_row} If ms\_network\_type is \texttt{OS\_MESH} the PEs are organized into rows and columns and this parameter is used together with ms\_col instead of ms\_size.
    
    \item \textbf{ms\_col} If ms\_network\_type is \texttt{OS\_MESH} the PEs are organized into rows and columns and this parameter is used together with ms\_row instead of ms\_size.
    
    \item \textbf{dn\_bw} and \textbf{rn\_bw} are the distribution and reduction bandwidth respectively as illustrated by Figure~\ref{fig:dnnacellstonne}.
    These parameters define the number of elements that can be distributed and reduced in a single cycle. If using the TPU, these must be specified as $\mathrm{dn\_bw}=\mathrm{ms\_rows}+\mathrm{ms\_cols}$ and $\mathrm{rn\_bw}=\mathrm{ms\_rows}*\mathrm{ms\_cols}$.
    \emph{Bifrost} enforces the TPU restriction and will correct improperly configured distribution and reduction networks.
   
    \item \textbf{reduce\_network\_type} is the type of reduction network.
    \texttt{ASNETWORK} refers to the ART reduction network proposed in MAERI~\cite{kwon2018maeri}, \texttt{FENETWORK} is an implementation of the STIFT reduction network~\cite{munoz-martinez2021}. 
    The TPU architecture must use the \texttt{TEMPORALRN} option. 

    \item \textbf{sparsity\_ratio}. This parameter defines the sparsity of the model. It is only used for the SIGMA architecture.
    
    \item \textbf{accumulation\_buffer}. Sets the accumulation buffer required to be enabled for rigid architectures like the TPU.
\end{enumerate}

\begin{table}[t]
\vspace{1.5mm}
    \caption{\emph{Bifrost}'s supported hardware configuration options for DNN accelerators in STONNE.}
    \label{tab:cfg}
    \centering
    \begin{tabular}{|c|c|}%
    \hline

    \textbf{Name}                & \textbf{Values}                \\
    \hline   
     \texttt{controller\_type} &  \makecell{\texttt{MAERI\_DENSE\_WORKLOAD}, \\ \texttt{SIGMA\_SPARSE\_GEMM}, \\ or \texttt{TPU\_OS\_DENSE}}\\
    \hline  \texttt{ms\_network\_type} & \texttt{OS\_MESH} or \texttt{LINEAR}  \\
    \hline \texttt{ms\_size}  &   $\{x \mid x \geq 8 \wedge \dfrac{log x}{log 2}\in  \mathbb{Z} \} $ \\ 
    \hline \texttt{ms\_row}  &   $\{x \mid \dfrac{log x}{log 2}\in  \mathbb{Z} \} $ \\ 
    \hline \texttt{ms\_col}  &   $\{x \mid \dfrac{log x}{log 2}\in  \mathbb{Z} \} $ \\ 
    \hline \texttt{dn\_bw} & $\{x \mid \dfrac{log x}{log 2}\in  \mathbb{Z} \} $ \\
    \hline \texttt{rn\_bw} & $\{x \mid \dfrac{log x}{log 2}\in  \mathbb{Z} \} $  \\
    \hline \texttt{reduce\_network\_type} & \makecell{\texttt{ASNETWORK}, \texttt{FENETWORK}, \\ or \texttt{TEMPORALRN}}   \\

    \hline
    \texttt{sparsity\_ratio} & $\{x \mid x \in  \mathbb{Z} \wedge 0 \leq x 100  \} $ \\
    \hline
    \texttt{accumulation\_buffer} & True or False   \\
    \hline     
    
    \end{tabular}
\end{table}

\section{Bifrost mapping optimization} 
\label{explore}

The AutoTVM module finds optimal hardware configurations and mappings for DNN models based on the cycle count or the count of partial sums required (\emph{psums}). 
This module leverages the tuners available in TVM such as grid search, GATuner~\cite{feng2010gatuner} (genetic algorithms), and XGBoost~\cite{chen2016} (a tree boosting system).

\subsection{Differences with standard AutoTVM} 

In typical usage of AutoTVM, the user searches the configuration space of a predefined schedule\footnote{A description of transformations and optimizations to be applied to a given algorithm on a target platform such as a CPU or GPU.} for a given operation.
For example, when running a conv2d layer on a CPU, TVM may define a schedule for the operation with loop reorderings and vectorization.
AutoTVM can search for parameters defining additional transformations which may improve performance, for instance the tile size to use and whether or not to unroll a given loop.
With this schedule and tuned parameters, TVM can then generate more efficient code used to run this operation.

This is in comparison to Bifrost's AutoTVM module, where the ``default schedule'' can be considered to be hardware accelerator design and AutoTVM searches for configuration parameters for the hardware accelerator.
An example of these parameters for MAERI is described in Section~\ref{subsec:dataflow}.

A recent alternative to AutoTVM is Ansor~\cite{zheng2020}, which searches for optimized schedules for CPUs and GPUs without the need for a default schedule constraining the search space.
This approach is called auto-scheduling and can significantly reduce the search time and potential performance improvement when compared to AutoTVM.
However auto-scheduling is not relevant to Bifrost, since we need to search for tuning parameters rather than schedules.
Ansor's dynamically trained predictive cost model may be valuable in reducing search costs, however we leave this exploration for future work.

\subsection{Optimization targets} 
\label{subsec:opt_targets}

The standard version of AutoTVM tunes schedules based on latency, i.e. the execution time of a layer.
As the latency will vary depending on many factors such as other system processes which may interfere with the result, the execution repeats several times per layer to find the average.
Latency is however not an appropriate optimization cost function when using STONNE. 
The latency of a layer executed on a simulated accelerator architecture using STONNE is not correlated with either the performance in terms of cycles nor other measurements of efficiency such as the simulated energy consumption.
For example, a simulated hardware architecture using more PEs will have a lower cycle count because of higher parallelism during execution but a higher latency, as PEs are not simulated in parallel in STONNE.
In other words, a faster simulation time does not mean that the simulated execution was faster.
Therefore, a custom cost function based on metrics reported from STONNE simulation is used instead.

\emph{Bifrost} can optimize performance targeting cycles or \emph{psums} (partial sums).
As STONNE is cycle-accurate both of these metrics are deterministic and multiple measurements are not needed.
Support for energy usage and area will be made available in Bifrost once STONNE has fully integrated them. 
When focused on reducing inference time, using cycle counts is the most accurate metric to optimize for, as the cost function is based on the reported cycles for a layer given a hardware configuration and mapping.
However, optimizing using cycles can be prohibitively slow for large models as the execution of a single layer can take many hours, and AutoTVM would need to run each layer many times with varying mappings.
To demonstrate how this could be problematic, the search space to generate an optimal mapping for a convolution simulated on the MAERI architecture where each tile has $10$ options would have $10^8$ (or $100$ million) possible combinations in the mapping space.
Exploring just $0.1\%$ of this mapping space would take around $347$ days if running the tuning process on a $12$ thread Intel Core i7 CPU in parallel and if each cycle count took $1$ hour to calculate. 
A cheaper alternative is to use \emph{psums} when tuning.
In this case, STONNE calculates the required amount of partial sums to execute the whole layer, a process that takes less than a second.
The \emph{psum} count can be used as a tuning value, which means that the layer does not have to be executed.

The intuition behind using \emph{psums} instead of cycles is that when fewer \emph{psums} are required the execution should be more efficient.
The amount of cycles to calculate each \emph{psum} does however vary, which means that using \emph{psums} for tuning is unlikely to generate the most optimal mapping. 
The trade-off is that exploring the same search space as in the previous example will take around $2$ hours when tuning using \emph{psums} instead of a year.

Note that this relationship is not necessarily linear, as the execution time will depend on many factors such as the configuration of the distribution network (i.e., bandwidth), the number of multipliers which affect the parallelism, and even the tile configuration which defines what partial sums are run in parallel. 
Thus the relationship between \emph{psums} and clock cycles is merely a correlation rather than strictly proportional.
Exploring the relationship between all of these factors is an ongoing area of research for reconfigurable DNN accelerators, which tools such as Bifrost will make easier to explore.

\subsection{MAERI dataflow configuration} 
\label{subsec:dataflow}

This section is primarily concerned with MAERI, as this is currently the only manually reconfigurable hardware architecture supported by \emph{Bifrost}.
SIGMA is also a flexible architecture, but the data flow mapping is automatically generated by the memory controller depending on the sparsity ratio. 
While only MAERI can use this module, support can be added when new architectures are added to STONNE.

If the user does not provide a mapping (tile pattern) a basic one will be generated. This means setting all tiles in Table~\ref{tab:maeri_conv} and~\ref{tab:maeri_fc} to $1$.
Execution using this mapping will be inefficient, but it makes it possible for researchers to quickly evaluate an architecture.
Section~\ref{explore} demonstrates how AutoTVM can be used to find optimal mappings.

\begin{table}[h]
    \caption{Mapping (tile) options for convolutions on MAERI~\cite{kwon2018maeri,stonne2021iiswc}.}
    \label{tab:maeri_conv}
    \centering
\begin{tabular}{|c|Q{0.84\linewidth}|}
\hline
\textbf{Tile} & \textbf{Description} \\
\hline
T\_R   & Number of filter rows mapped at a time                                    \\ \hline
T\_S   & Number of filter columns mapped at a time                                 \\ \hline
T\_C   & Number of filter \& input channels per group mapped at a time            \\ \hline
T\_K   & Number of filter \& output channels per group mapped at a time          \\ \hline
T\_G   & Number of groups mapped at a time                                         \\ \hline
T\_N   & Number of inputs mapped at a time (STONNE supports only 1)  \\ \hline
T\_X   & Number of output rows mapped at a time                                 \\ \hline
T\_Y   & Number of input columns mapped a time \\
\hline
\end{tabular}
\end{table}

\begin{table}[h]
    \caption{Mapping (tile) for fully connected layers on MAERI~\cite{kwon2018maeri,stonne2021iiswc}.}
    \label{tab:maeri_fc}
        \centering
\begin{tabular}{|c|c|}
\hline
\textbf{Tile} & \textbf{Description} \\
\hline
T\_S & Number of output neurons mapped at a time \\ \hline
T\_N & Number of batches mapped at a time \\ \hline
T\_K & Number of input neurons mapped at a time \\
\hline
\end{tabular}
\end{table}

\subsection{Specialized Mapping Tools}
\label{subsub:special_map_tools}

Bifrost's AutoTVM module represents a valuable contribution in providing a simple accelerator architecture-agnostic tool that searches for optimized configuration parameters. 
Our evaluation in Section~\ref{subsec:eval_mapping} shows that it can generate competitive mappings.
However, since it assumes no knowledge of the underlying hardware, relying on metrics generated from STONNE can mean that its search strategies are suboptimal.
Specialized mapping tools that encode the features of the reconfigurable accelerator architecture may be able to find more optimal mappings in less time.
However, these tools must be created often with a high engineering cost.

Nevertheless, when these tools are available Bifrost has a mechanism to integrate and exploit them.
In the case of the MAERI architecture, the mRNA tool~\cite{zhao2019mrna} achieves this goal and Bifrost can use it to generate mappings.
Thus, Bifrost can automatically produce optimized mappings both in settings where a specialized mapping tool is not available and where one is available.
In comparison, STONNE only works in the latter case, otherwise the mapping must be generated by hand.

\section{Evaluation} \label{eval}

The evaluation of \emph{Bifrost} is divided into two parts. 
The first part illustrates how \emph{Bifrost} can be used to evaluate the performance of a given DNN model on different accelerator architecture configurations. 
The second part discusses how Bifrost's AutoTVM module can be used to generate an optimized mapping and how this mapping compares to other expert tools (such as mRNA) also integrated into Bifrost.
Throughout this evaluation AlexNet~\cite{krizhevsky2012imagenet}, a canonical convolutional neural network (CNN), is used to provide layers for benchmarking.
Only 2D convolutional and fully connected layers are evaluated, as these are the only layer types currently supported by STONNE.
However, extensions to STONNE can be easily supported by Bifrost, such as adding support for new layers, and power and area metrics.

Hardware optimizations are not evaluated (i.e., varying the amount of hardware resources available rather than their configuration).
The hardware configuration options which dictate the amount of available resources to a simulated accelerator correlate strongly with the clock cycles, and thus we do not need to use the optimization module to confirm that more powerful (simulated) hardware is indeed faster, our experiment shown in Figure~\ref{fig:conv2d} shows this implicitly.
However, evaluating the impact of optimizing hardware parameters becomes relevant if the target is investigating the trade-off between performance and energy consumption, or area. 
STONNE's extension to support energy and area metrics is still under development.
When available \emph{Bifrost} will be able to add them as optimization targets, thus making this evaluation relevant.

\subsection{Comparing architecture configurations}

\begin{figure}[t]
  \centering
     \centering
     \begin{subfigure}[b]{0.465\linewidth}
         \centering
         \includegraphics[width=\linewidth]{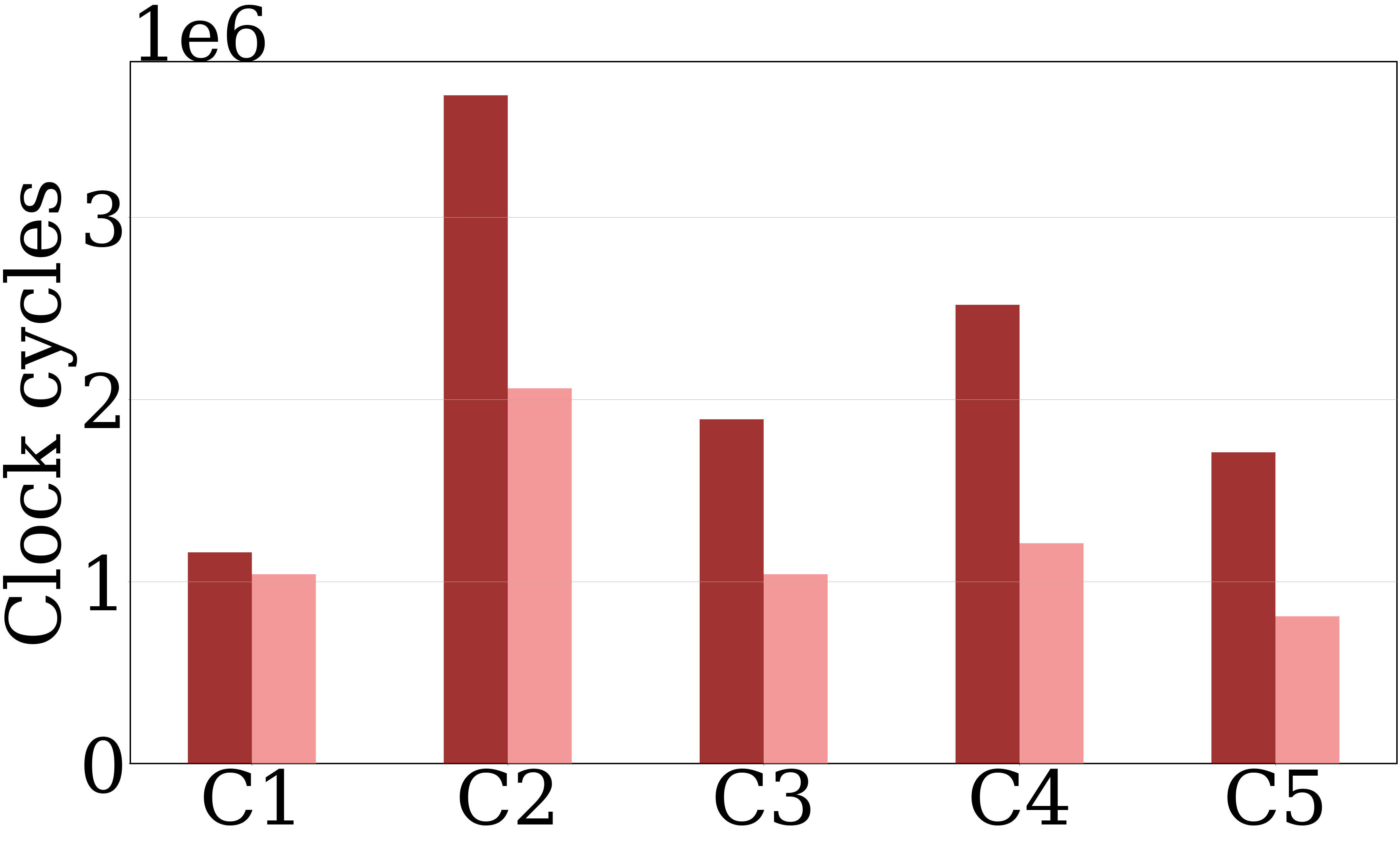}
         \caption{Convolutional layers}   \centering
    \end{subfigure}
    \hfill
    \begin{subfigure}[b]{0.465\linewidth}
         \centering
         \includegraphics[width=\linewidth]{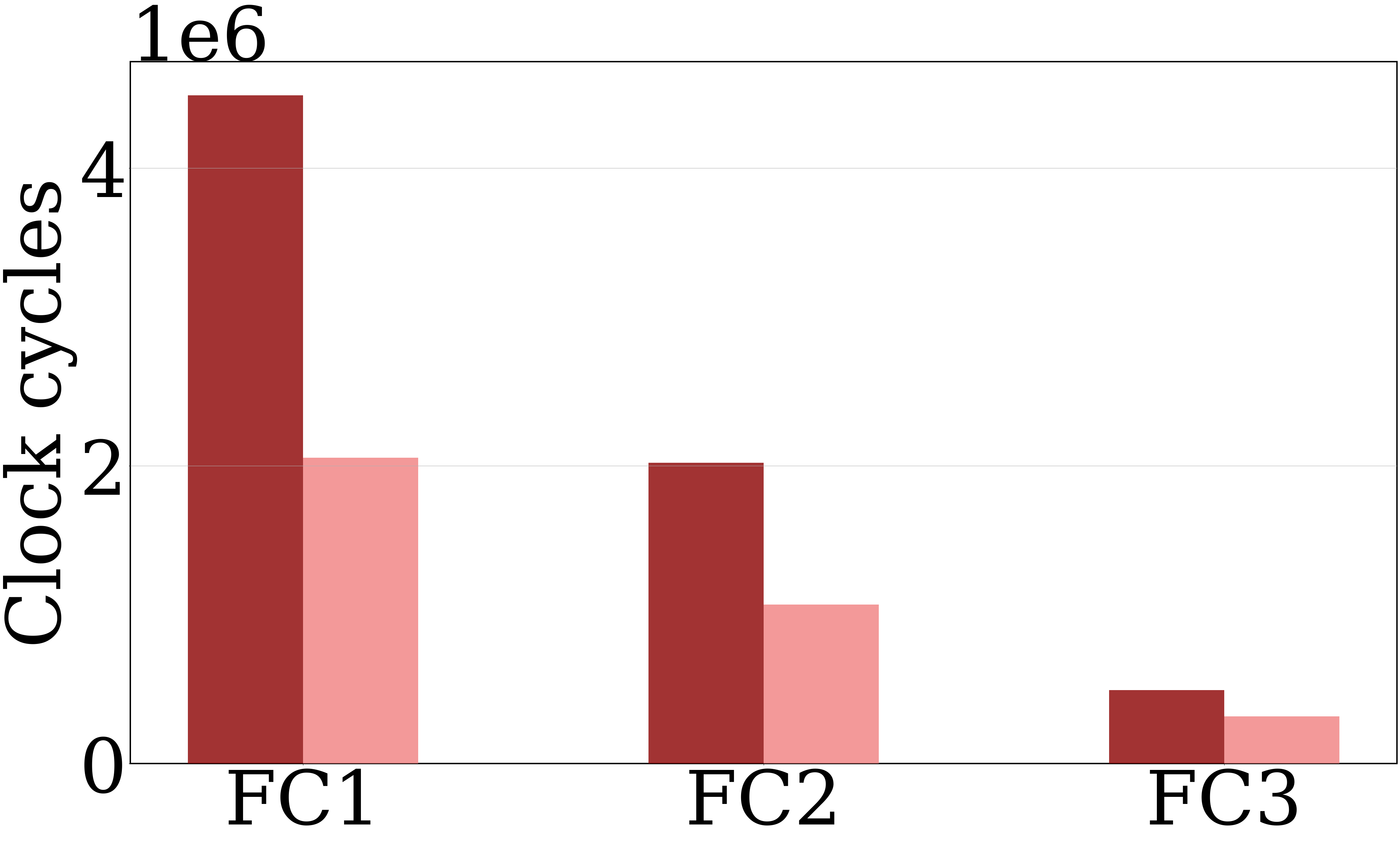}
         \caption{Fully connected layers}   \centering
    \end{subfigure}
 
  \caption{
Clock cycles for 0\% and 50\% sparsity for the convolutional (a) and fully connected (b) layers in AlexNet running on simulated SIGMA architecture.}
  \label{fig:alexnet_sigma}
\end{figure}

\emph{Bifrost} can be used for quickly evaluating the inference performance at different architecture configurations.
For example, when evaluating the SIGMA architecture the dataflow orchestration is automatically handled by the memory controller depending on the sparsity setting. 
Therefore the performance in terms of cycles is dependent on the sparsity setting. 
Figure~\ref{fig:alexnet_sigma} shows AlexNet executed using SIGMA simulated on STONNE with different levels of pruning.
The results are roughly in line with what would be expected. On average, the convolutional layers require $44\%$ fewer cycles and the fully connected layers require $54\%$ fewer cycles when running using sparsity set at $50\%$.
While outside the scope of this evaluation, this kind of architecture comparison can be used to find the trade-off between the accuracy of the model at different levels of sparsity to the performance when executed using SIGMA.

The example above demonstrates how \emph{Bifrost} can be used to evaluate the performance of different architecture configurations for a given DNN model and how these results can be incorporated into further research.
Using \emph{Bifrost}, researchers can effortlessly evaluate any hardware parameter configuration and find how tweaking them affects the performance.

\subsection{Mapping optimization}
\label{subsec:eval_mapping}

\begin{figure}[t]
  \centering
  \includegraphics[width=0.85\linewidth]{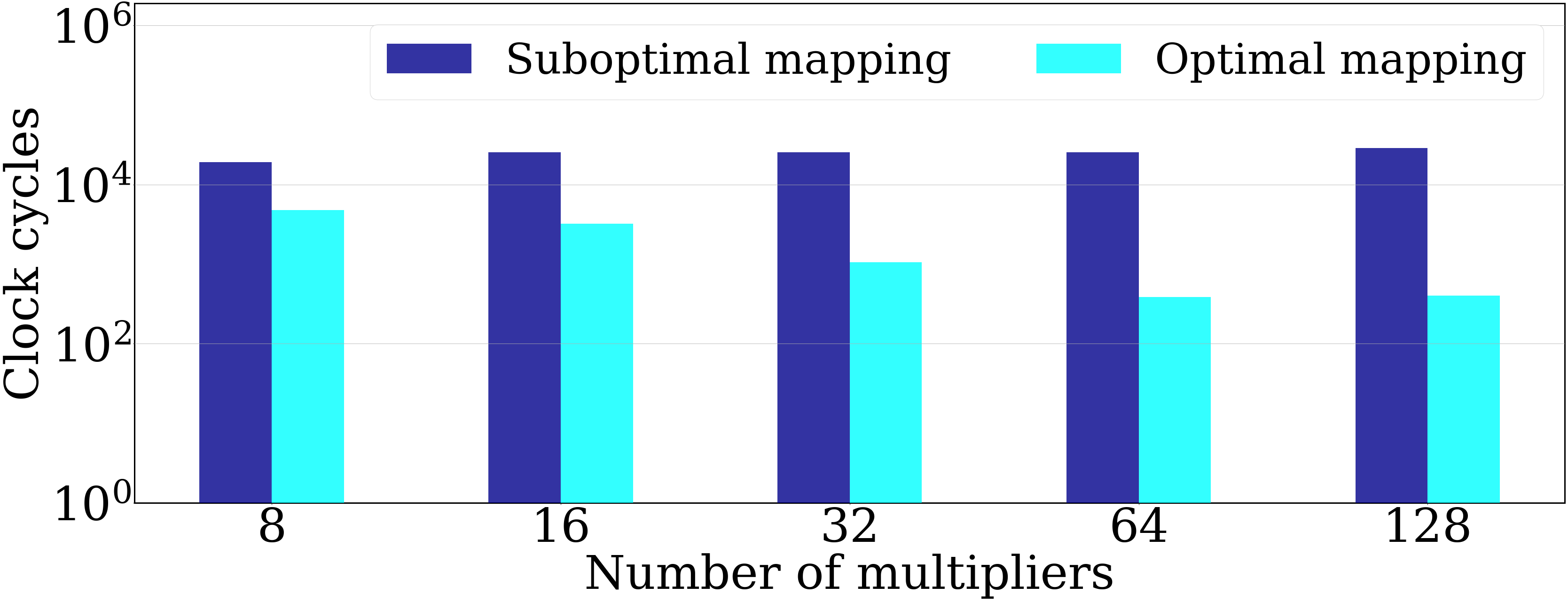}
  \caption{An NCHW convolution in MAERI with $1 \times 2 \times 10 \times 10$ input dimensions, executed with optimal and suboptimal mappings. The scale of the y-axis is logarithmic.
  }
  \label{fig:conv2d}
\end{figure}

Finding an optimal dataflow mapping is crucial for reconfigurable accelerators such as MAERI.
Figure~\ref{fig:conv2d} demonstrates the impact of dataflow mappings by running a small \texttt{NCHW} convolution ($1 \times 2 \times 10 \times 10$ input tensor with random data) simulated on MAERI.
The same convolution was executed using an increasing number of multipliers (i.e., PEs or multiply-accumulate units), thus increased hardware resources. 
For each multiplier setting, mappings were generated using \emph{Bifrost}'s AutoTVM module optimizing for cycles using an exhaustive grid-search over the whole mapping space.
From this mapping space the \emph{suboptimal} (worst) and \emph{optimal} (best) mappings have been selected for evaluation. 
This exhaustive grid search ensures that we find the globally optimal and suboptimal mappings, however in reality this is too expensive, thus we should use tuners like XGBoost~\cite{chen2016} to more efficiently search a subset of mapping space.

This experiment clearly shows the impact of dataflow orchestration. 
The whole mapping space is searched for each multiplier setting.
For small accelerator architectures (i.e., few PEs), the clock cycle count for the suboptimal mapping and the optimal mapping differ by a factor of around $4$.
As expected, when using an optimal mapping the amount of multipliers is inversely correlated with the amount of clock cycles required to execute the workload. 
When using optimal dataflow mappings, $8$ multipliers require about $12$ times more clock cycles to compute the same workload as $128$ multipliers. 
However, the suboptimal mappings perform increasingly worse with the amount of multipliers, which demonstrates the impact and importance of dataflow orchestration for larger and more complex architectures.
When executing the convolution with $128$ multipliers, the required clock cycles for the optimal and suboptimal mappings differ by a factor of $76$.
Reconfigurable accelerators like MAERI are able to efficiently execute DNN workloads, but only if provided with efficient mappings from a tool like \emph{Bifrost}'s AutoTVM module.

\begin{figure}[t]
  \centering
     \centering
     \begin{subfigure}[b]{0.465\linewidth}
         \centering
         \includegraphics[width=\linewidth]{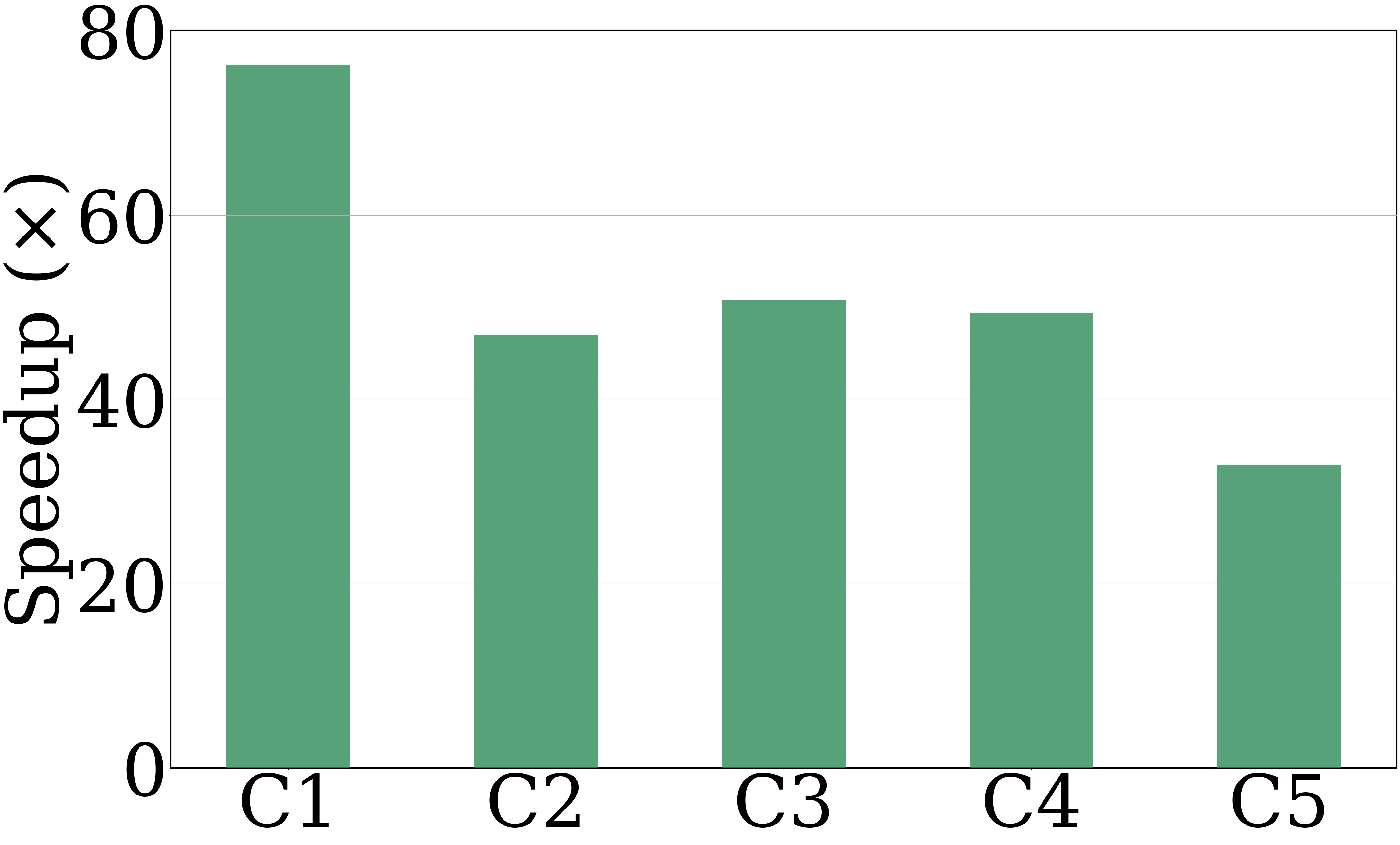}
         \caption{Convolutional layers \label{fig:maeri_conv}}   \centering
    \end{subfigure}
    \hfill
    \begin{subfigure}[b]{0.465\linewidth}
         \centering
         \includegraphics[width=\linewidth]{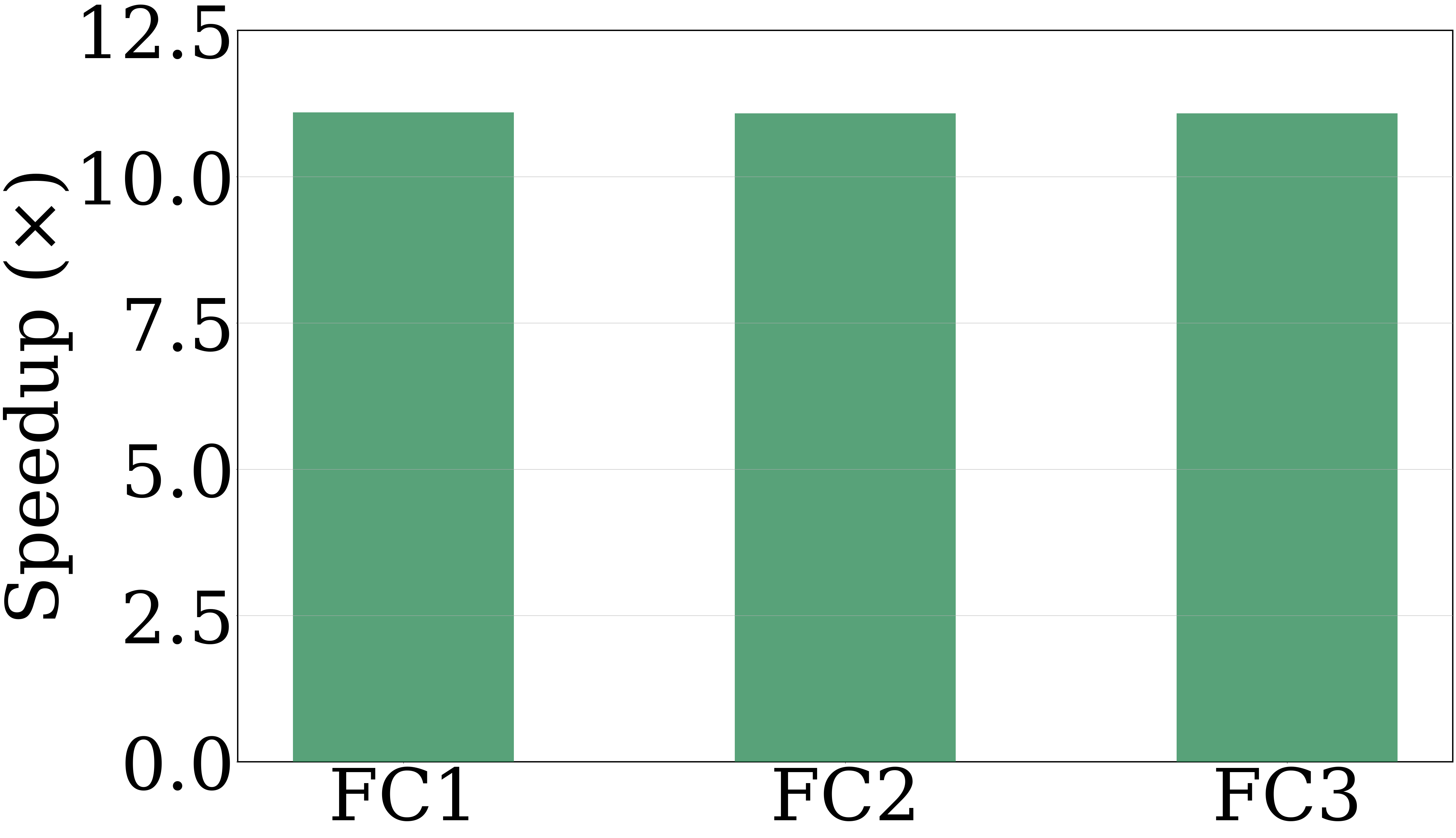}
         \caption{Fully connected layers \label{fig:maeri_fc}}   \centering
    \end{subfigure}
  \caption{Performance speedup for the convolutional (a) and fully connected (b) layers in AlexNet running on simulated MAERI with 128 multipliers. 
  An optimized dataflow mapping for each layer has been generated with \emph{Bifrost}.}
  \label{fig:alexnet_maeri_speedup}
\end{figure}

To demonstrate the mapping optimization process on a wider range of layer sizes and types, we evaluate the 5 conv2d and 3 fully-connected layers of AlexNet using the MAERI architecture. 
MAERI requires a mapping to be provided when executing a workload and \emph{Bifrost} will automatically generate an unoptimized default mapping if none is provided.
Rather than use the exhaustive grid search described above, we take the default mapping as our baseline, and compare against an optimized mapping (which may not be the global optimum).
We use \emph{psums} as the metric to optimize for (as tuning based on cycles would take weeks), with XGBoost as the tuner, stopping once we reach convergence.
This is achieved using AutoTVM's ``early stopping'' utility.

Figure~\ref{fig:maeri_conv} illustrates the performance speedup of the convolutional layers when using the efficient mapping for each layer generated by AutoTVM, compared to the basic mappings. 
On average a $51\times$ speedup is demonstrated, with a maximum speedup of $77\times$. 
Figure~\ref{fig:maeri_fc} shows the performance speedup when using the AutoTVM optimized mapping for the fully connected layers. 
While the speedup is smaller than the convolutional layers, the optimized mappings demonstrate an average speedup of $11\times$ compared to the default mappings, with all layers seeing a similar speedup.

Observing the generated mapping for these layers obtained from the AutoTVM module,  Table~\ref{tab:fc_mapping} gives a hint as to why these mappings are sub-optimal. 
The table shows that the AutoTVM module always maximizes the \emph{T\_S} tile (number of output neurons mapped) while always minimizing \emph{T\_N} (number of batches mapped) and \emph{T\_K} (number of input neurons) when the optimization target is minimizing psums.
The AutoTVM mappings optimizing for \emph{psums} are not able to achieve globally optimal dataflow orchestration, as they are not able to vary the mapping depending on the layer characteristics.

\begin{table}[h]
    \caption{Mappings for the fully connected layers in AlexNet executed using simulated MAERI, comparing the basic mapping with the mappings generated by AutoTVM and mRNA. The tiles for fully connected layers in MAERI are T\_S, T\_K, T\_N~\cite{kwon2018maeri}, see Table~\ref{tab:maeri_fc} for full descriptions.}
    \label{tab:fc_mapping}
    \begin{tabularx}{\linewidth}{|C|C|C|C|}
    \hline
    \textbf{Mapping} & \textbf{FC1 } & \textbf{FC2} & \textbf{FC3} \\  
    \hline
    Basic & 1, 1, 1 & 1, 1, 1 & 1, 1, 1 \\
    AutoTVM & 20, 1, 1 & 20, 1, 1  & 20, 1, 1 \\
    mRNA & 12, 8, 1 & 16, 4, 1 & 8, 10, 1\\
    \hline     
    \end{tabularx}
\end{table}

Comparing our AutoTVM mappings with those chosen by mRNA, the latter can vary the size of \emph{T\_N} and \emph{T\_K} for each layer for optimal dataflow orchestration.
mRNA performs better as it explicitly encodes the design of the MAERI architecture, and thus can make more informed choices.
mRNA uses domain knowledge about MAERI to generate an efficient dataflow mapping, while AutoTVM optimizes the dataflow purely based on metrics from iterative simulations.
Additionally, mRNA is more efficient taking minutes rather than hours to produce its mappings, since it does not need to run a simulation.
Thus, mRNA mappings were generated with the goal of minimizing the total cycle count for each layer, compared to AutoTVM where we used \emph{psums} to reduce the search time (as discussed in Section~\ref{subsec:opt_targets}).

\begin{figure}[ht]
  \centering
     \centering
     \begin{subfigure}[b]{0.465\linewidth}
         \centering
         \includegraphics[width=\linewidth]{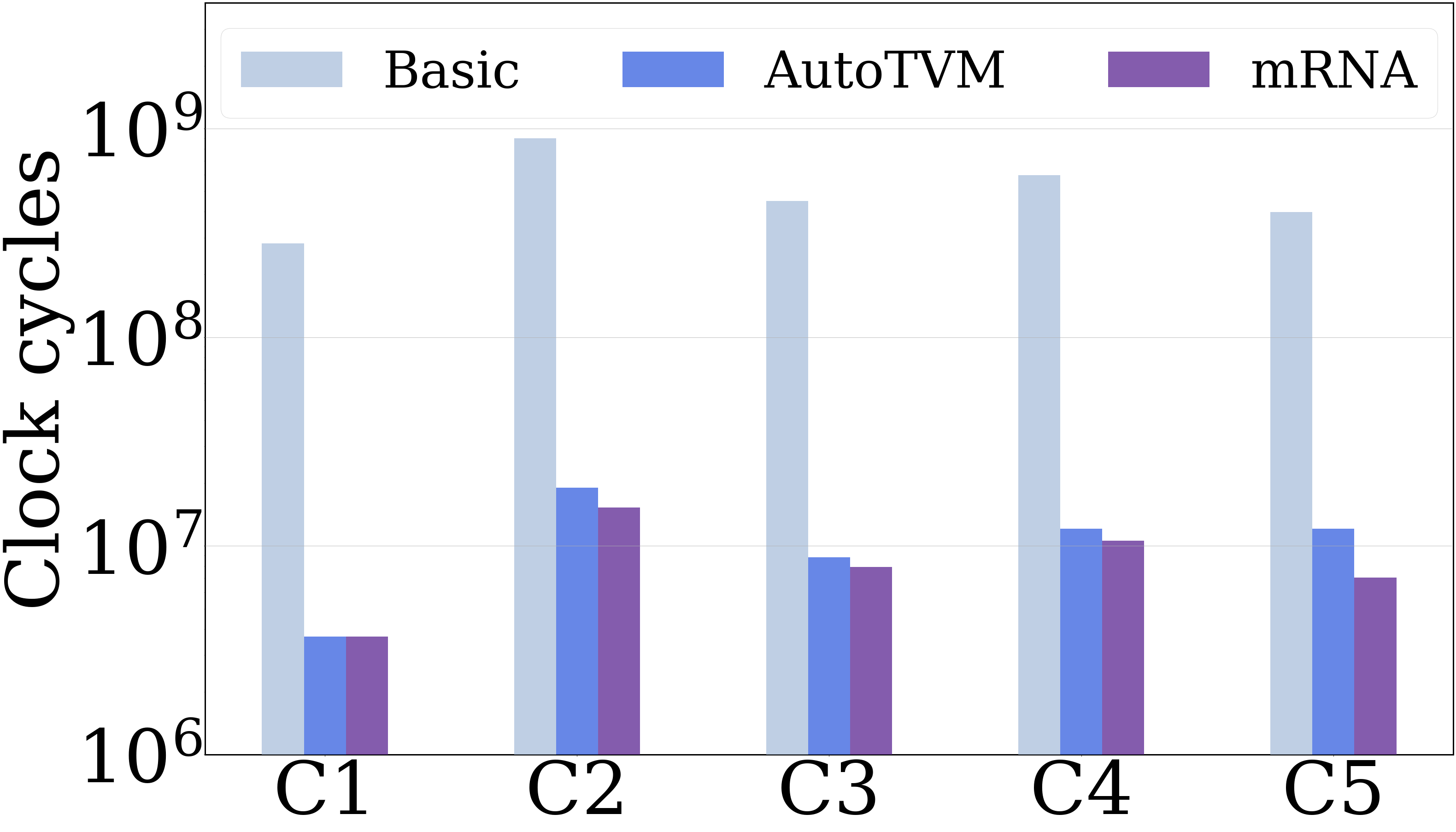}
         \caption{Convolutional layers \label{fig:mappings_conv}}   \centering
    \end{subfigure}
    \hfill
    \begin{subfigure}[b]{0.465\linewidth}
         \centering
         \includegraphics[width=\linewidth]{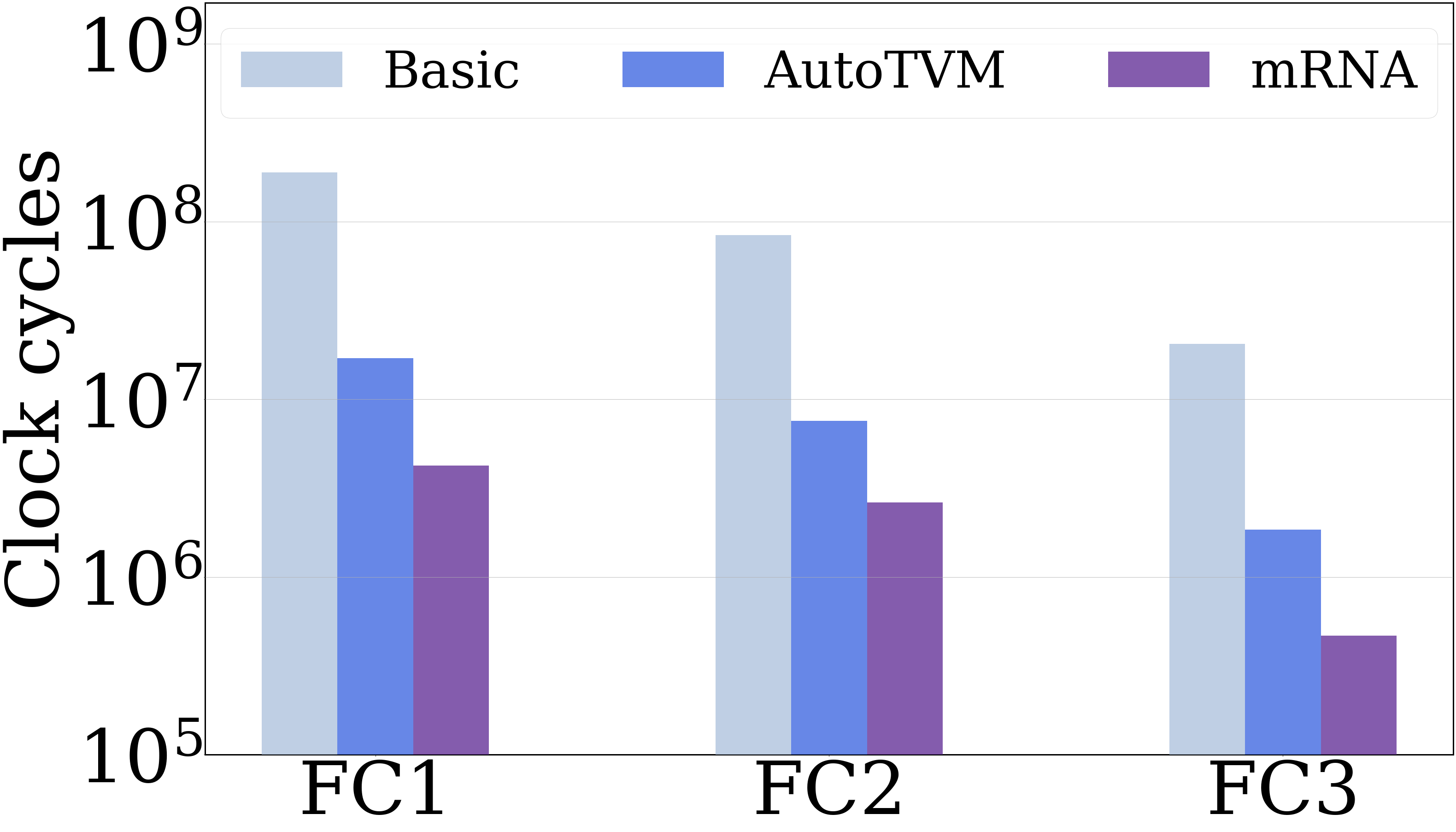}
         \caption{Fully connected layers \label{fig:mappings_fc}}   \centering
    \end{subfigure}
  \caption{Clock cycles using different mappings for AlexNet's convolutional (a) and dense (b) layers running on simulated MAERI architecture. The y-axis scale is logarithmic.
  }
  \label{fig:alexnet_maeri_cycles}
\end{figure}

To show how well Bifrost's AutoTVM module compares to expert mapping systems (such as mRNA), the cycle counts for all layers of AlexNet using the dataflow mappings generated by STONNE by default, AutoTVM, and mRNA are shown in Figure~\ref{fig:alexnet_maeri_cycles}. 
The figure shows how the \emph{psums} count is merely loosely correlated with performance.
Optimizing based on \emph{psums} produces efficient mappings but not optimal ones.
This works reasonably well for convolutional layers but not for fully connected layers.
This shows that Bifrost+AutoTVM is fit for purpose when evaluating a reconfigurable accelerator design that does not have an optimized mapping tool such as mRNA, but will not necessarily find the optimal configuration.
This could make it valuable during accelerator design exploration.

Figure~\ref{fig:mappings_conv} shows how the mapping generated by mRNA requires on average $20\%$ fewer cycles than the one generated by AutoTVM. 
A similar trend can be observed for the fully connected layers in Figure~\ref{fig:mappings_fc}, where the mRNA mapping requires on average $67\%$ fewer cycles compared to the AutoTVM mapping.

An important observation is that AutoTVM can perform almost as well as expert tools (e.g., mRNA) while assuming no knowledge about the underlying architecture.
AutoTVM also tuned the dataflow based on the \emph{psums} count which is only loosely correlated to performance.
More efficient mappings could most likely be obtained by tuning using cycle counts, however AutoTVM is still limited by the execution time of STONNE, so this would take a prohibitively long time.
These results show that Bifrost+AutoTVM can produce mappings with similar efficiency compared to expert systems such as mRNA, and would be ideal for other novel reconfigurable architectures with no expert tools available to optimize the hardware.
For example, this could be valuable during the development of novel reconfigurable accelerator designs.
Bespoke mapping tools should be considered an end goal of mature reconfigurable accelerator design, and can be integrated into Bifrost when available.

\section{Conclusion}    

This paper presented \emph{Bifrost}, a tool built on STONNE~\cite{stonne2021iiswc} and Apache TVM~\cite{chen2018tvm} that enables accessible end-to-end evaluation and optimization of reconfigurable DNN accelerators. 
The main challenges of using STONNE were identified, as the limited support for different deep learning frameworks and the significant manual effort required to create architecture configuration and mapping files. 
To address these challenges we connected STONNE with TVM, with TVM's support of a wide range of deep learning frameworks solving the first issue, and its learning-based cost model AutoTVM being used to explore architecture design and dataflow mapping space.

We evaluated Bifrost on the SIGMA architecture at varying levels of sparsity. 
For the MAERI architecture, which requires user defined reconfiguration, we compared using the mRNA mapping tool (integrated with Bifrost) against AutoTVM to generate mappings.
The mappings identified by AutoTVM required only $20\%$ more clock cycles for the convolutional layers and $67\%$ more for the fully connected layers. 
This makes \emph{Bifrost+AutoTVM} potentially suitable for optimizing novel reconfigurable architecture designs which do not have yet have expert tools available.

As future work, we would like to extend Bifrost to support AutoTVM tuning using other optimization targets such as energy efficiency and add support for more operators such as sparse-dense matrix multiplication~\cite{nichols2019magmadnn}, which would allow other accelerator designs like MAGMA~\cite{nichols2019magmadnn} to be evaluated.

\balance
\bibliographystyle{IEEEtran}
\bibliography{IEEEabrv,bibfrost}

\end{document}